# Recent Advances in Generative AI for Healthcare Applications

Yasin Shokrollahi[1], Jose Colmenarez[1], Wenxi Liu[2], Sahar Yarmohammadtoosky[3], Matthew M. Nikahd[4], Pengfei Dong[1], Xianqi Li[2*], Linxia Gu[1*]

[1] Department of Biomedical Engineering and Sciences, Florida Institute of Technology, Melbourne, FL 32901, USA
[2] Department of Mathematics and Systems Engineering, Florida Institute of Technology, Melbourne, FL 32901, USA
[3] Department of Ira A. Fulton Schools of Engineering, Arizona State University, Temple, AZ 85281, USA
[4] Department of Data Science and Analytics, Kennesaw State University, Kennesaw, GA 30144, USA
* Corresponding authors: xli@fit.edu, gul@fit.edu

## Abstract

The rapid advancement of Artificial Intelligence (AI) has catalyzed revolutionary changes across various sectors, notably in healthcare. In particular, generative AI-led by diffusion models and transformer architectures-has enabled significant breakthroughs in medical imaging (including image reconstruction, image-to-image translation, generation, and classification), protein structure prediction, clinical documentation, diagnostic assistance, radiology interpretation, clinical decision support, medical coding, and billing, as well as drug design and molecular representation. These innovations have enhanced clinical diagnosis, data reconstruction, and drug synthesis. This review paper aims to offer a comprehensive synthesis of recent advances in healthcare applications of generative AI, with an emphasis on diffusion and transformer models. Moreover, we discuss current capabilities, highlight existing limitations, and outline promising research directions to address emerging challenges. Serving as both a reference for researchers and a guide for practitioners, this work offers an integrated view of the state of the art, its impact on healthcare, and its future potential.

## 1. Introduction:

Generative models began their journey in the 1950s and have seen major improvements in the last decade, changing how we understand deep learning. In the early days, concepts like Hidden Markov and Gaussian Mixture models introduced simple data generation methods. However, the integration of neural networks revolutionized these methods, leading to substantial improvements in representational capacity and generalization.

Today, generative artificial intelligence (AI) has demonstrated remarkable capabilities across diverse data modalities (**Figure 1**). In natural language processing (NLP), models such as OpenAI's GPT series (Brown et al. 2020a), have proven to be a powerful tool for generating detailed and logical stories from text inputs. In computer vision, generative AI models have been specifically designed to handle image data, as referenced in (Ramesh et al. 2021), for purposes such as image generation and resolution enhancement, while in video generation, temporal modeling allows realistic motion synthesis (Ho et al. 2022). Moving beyond just 2D media, generative AI has made the creation of 3D objects and spatial landscapes possible, leading to revolutionary developments in virtual reality and video games, as indicated by (Shen et al. 2021).

The continual enhancement of AI models and their expanding capability to handle various types of data have significantly impacted the field of healthcare. It is now being used in applications including MRI image quality enhancement (Özbey et al. 2023), cross-modality image translation (Isola et al. 2017), synthetic medical dataset generation (Müller-Franzes et al. 2023), 3D anatomical reconstruction from segmented images (Pinaya et al. 2022), protein structures prediction (Madani et al. 2023), and numerous other diagnostic and therapeutic tasks. This entire journey and evolution of generative models have been documented and studied by many researchers, including (Bao et al. 2017; Razavi, Van den Oord, and Vinyals 2019; Kong et al. 2020; Oord et al. 2016; Li et al. 2022; Yang et al. 2019). In particular, recent advances in diffusion models have further expanded these capabilities: multi-conditioned denoising diffusion probability models (DDPMs) for lung CT synthesis (Krishna et al., 2025), semantic layout–guided 3D thoracic CT generation (Jiang et al., 2025), and resource-efficient high-resolution 3D CT synthesis via MedLoRD (Seyfarth et al., 2025) all demonstrate substantial gains in anatomical fidelity and scalability. Diffusion-based augmentations have also been shown to improve fairness and robustness in histopathology, chest X-ray, and dermatology classification (Ktena, 2024).

Modern generative AI owes much of its progress to deep learning architectures, including Generative Adversarial Networks (GANs) (Goodfellow et al. 2014), Variational Autoencoders (VAEs) (Rezende, Mohamed, and Wierstra 2014), normalizing flows (Dinh, Sohl-Dickstein, and Bengio 2016), and NeRF (Neural Radiance Fields) (Mildenhall et al. 2021). More recently, diffusion models (Ho, Jain, and Abbeel 2020) have emerged as a robust alternative, addressing limitations in VAEs, GANs, and energy-based models. Transformers, originally introduced in NLP tasks (Vaswani et al. 2017), have also transformed generative AI through self- and cross-attention mechanisms, surpassing recurrent models such as Recurrent Neural Networks (RNNs) (Rumelhart, Hinton, and Williams 1985), Long Short-Term Memory networks (LSTMs) (Hochreiter and Schmidhuber 1997), and Gated Recurrent Units (GRUs) (Chung et al. 2014) ( **Figure 2**). While not generative in the traditional probabilistic sense, transformers enhance or enable generation in hybrid systems. For example, SwinUNetR (Hatamizadeh et al. 2021) incorporates transformers architecture to achieve competitive results in tasks such as medical image segmentation. Transformers-based models have also seen expanding use in healthcare for text summarization, named entity recognition, multimodal fusion, and dialogue systems (Li and Tang, 2024; Liu, et al., 2025). The interplay between transformers and other generative models highlights the importance of considering them not only as standalone models but also as complementary tools.

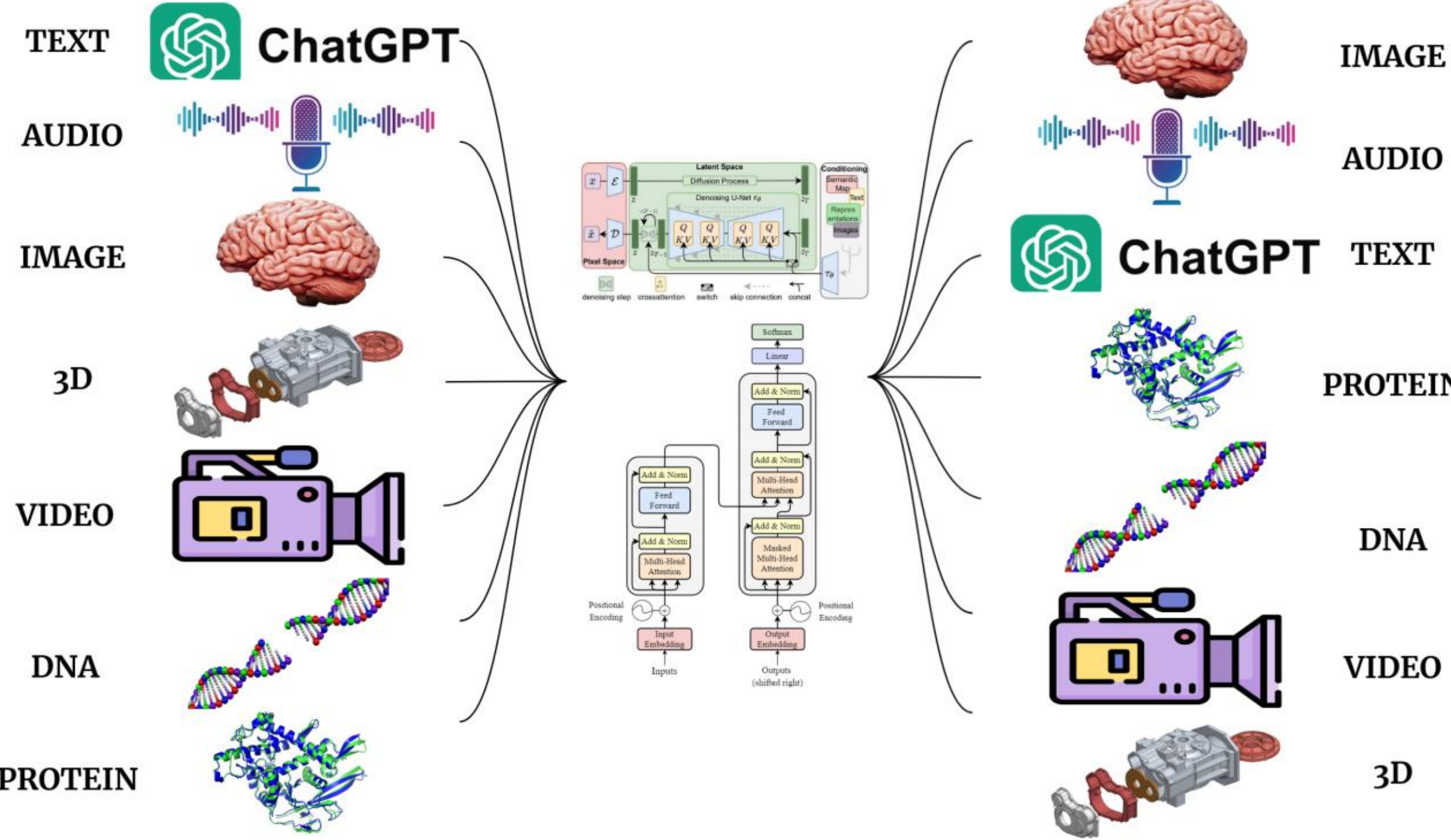

**Figure 1**. Overview of Generative AI tasks.

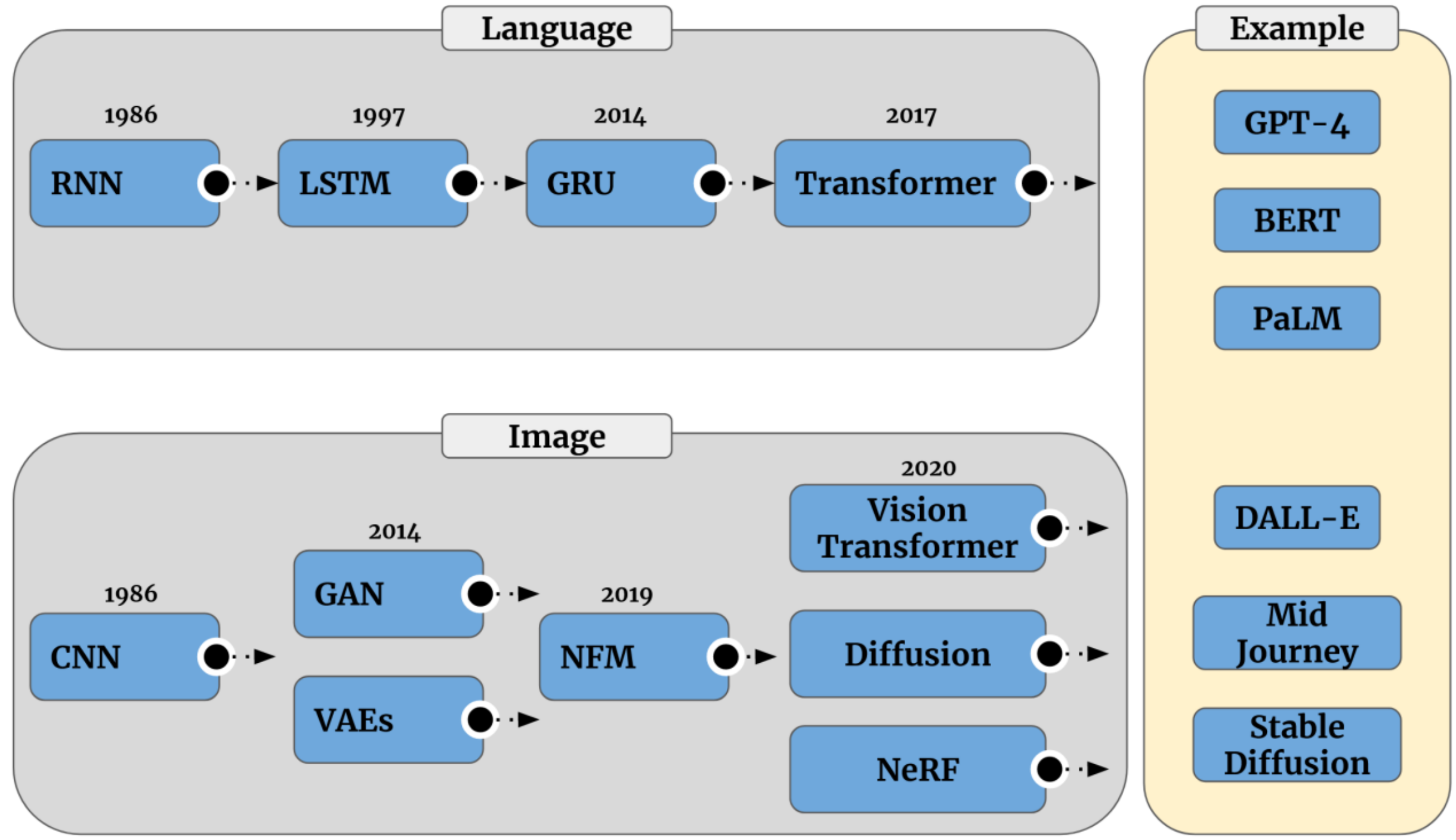

**Figure 2.** Timeline of Generative AI Family Types.

Since diffusion and transformer-based models have outperformed other types of models/architectures and are increasingly used in the healthcare industry (**Figure 3**), we have primarily opted to focus on these two in the manuscript. Our review extensively examines the latest applications of generative AI in healthcare, representing, to the best of the author's knowledge, the first thorough study of these AI techniques in this sector. The main contributions of this paper are as follows:

- A pioneering review that extensively captures the application of generative AI in healthcare, encapsulating a comprehensive assessment of all pertinent studies up to mid-2025 .
- We present a clear classification of generative AI models in healthcare and divide them into two main types: diffusion models and transformer-based models. We further categorize their uses: diffusion models can complete tasks like image reconstruction, image-to-image translation, image generation, image classification, and other applications, while transformer-based models have been used for protein structure prediction, clinical documentation and information extraction, diagnostic assistance, medical imaging and radiology interpretation, clinical decision support, medical coding and billing, and drug design and molecular representation.
- Our focus is not solely on applications. We propose a novel classification wherein each study is broadly sorted based on its underlying algorithm and imaging methodology.
- AI types of models have been compared in various applications, with the pros and cons of each model being discussed. For each specific task in healthcare, the best model based on research and publications has been examined.

In summary, we addressed the looming challenges and potential issues and identified emerging trends for the healthcare sector. We posed thought-provoking questions regarding the future direction of generative

AI in healthcare, covering both algorithms and practical considerations. Moreover, we believe that generative AI has the potential to be recognized as a highly reliable assistant for medical professionals in the future, and its use could become widespread in healthcare.

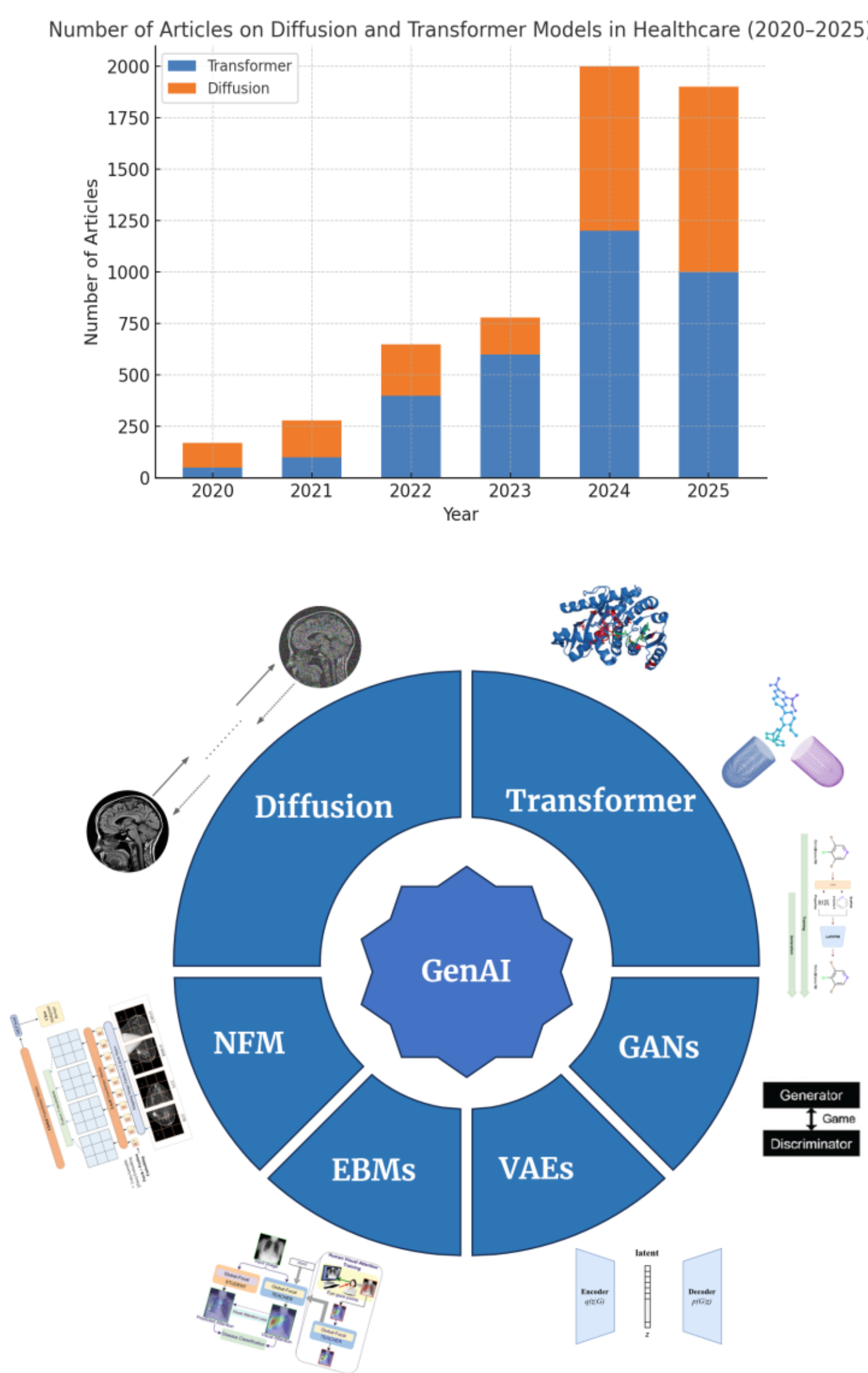

Figure 3. Generative AI in healthcare.

## 1.1. Rationale and Distinctiveness of Our Review:

In healthcare, generative AI has made significant progress over the years. Many experts have written detailed reviews about deep generative AI models designed specifically for healthcare purposes. These reviews include the studies by (Bohr and Memarzadeh 2020), (AlAmir and AlGhamdi 2022), (Ali et al. 2022), (Kazerouni et al. 2023), (Nerella, et al. 2024), Some of these reviews focus on specific uses or applications, while others explore different types of images produced by these models. Even though there have been previous reviews, numerous new advancements in healthcare have emerged since their publication.

Interestingly, there is a clear gap in the current reviews: they need to detail how generative AI is used in healthcare. This is a crucial topic because understanding this can push research in this area forward.

Much potential in this field has not been fully explored yet. The healthcare community can get valuable insights by understanding the benefits and results of using generative AI, especially in producing images and large language models (LLM).

This review examines past research comprehensively and what might be coming up regarding generative AI models in healthcare. We want to offer a complete viewpoint on this subject. Generative AI technology has shown great skill in creating artificial data. It is becoming a strong addition to our usual data and has a special role in some medical areas, which we will discuss in the next sections. With this paper, we want to guide medical experts in understanding and using these modern methods in their work. We look at these generative models and point out studies and writings that show how they are used in healthcare. However, we do not just stop explaining their use; we also explore how they work, the specific body parts they target, and the type of healthcare they relate to. By providing this deeper understanding, we hope to help researchers see how different studies and information connect, forming a clear story.

### 1.2. Search Methodology:

We searched several online platforms, namely DBLP, Google Scholar, PubMed, and Arxiv Sanity Preserver, using specific search queries designed for our study. These platforms allow for detailed search terms and provide lists of academic works. The types of content exhibited by these platforms include those from peer-reviewed journals, conference or workshop proceedings, articles that are not peer-reviewed, and early versions of papers. For our search, we used a set of terms related to advanced AI techniques and their healthcare applications. Specifically, we used search terms like (Generative AI* | transformer* | diffusion model*) (healthcare* | medicine* | medical* | medical imaging* | clinical* | diagnosis* | treatment*). After getting our search results, we filtered to ensure we only kept articles directly related to using generative AI models in healthcare. During our search, we came across many articles that, despite having relevant terms, were unrelated to our study. We did not include these in our final selection. When choosing which papers to focus on more deeply, we looked for those that were innovative, made a major contribution to the field, were highly regarded, and set a precedent in the healthcare arena. After being thorough in our selection criteria, we chose a handful – specifically, two or three of the best papers – for a detailed review. However, it is important to mention that even though we aimed to give a comprehensive overview of the most important articles on this topic, there is a chance that we might have missed a few significant papers during our review.

## 2. Generative AI Models:

As an advanced subclass of generative models, diffusion models have garnered significant attention for their adeptness at learning intricate data distributions. Their advancement in the generative learning domain is recent, yet their utility is evident across various applications. The forthcoming section thoroughly explores the theoretical aspects underpinning diffusion models. We initiated this discussion by situating diffusion models within the wider context of generative learning, offering fresh comparative insight against other generative models. Subsequently, we categorize diffusion models from two principal vantage points: **the Variational and Score perspectives**. Taking a closer look at each perspective, we provide insights into specific models classified under them. For instance, within the Variational Perspective, Denoising Diffusion Probabilistic Models (DDPMs) are notable, while Noise Conditional Score Networks (NCSNs) and Stochastic Differential Equations (SDEs) prominently feature within the Score Perspective.

With a surge in accessible datasets and advancements in deep learning architectures, there's been a transformative shift in generative modeling. Specifically, the three predominant generative frameworks, GANs (Goodfellow et al. 2014), VAEs (Rezende, Mohamed, and Wierstra 2014; Kingma and Welling 2013), and normalizing flows (Dinh, Sohl-Dickstein, and Bengio 2016), have come to the fore (refer to Fig. 2). Adoption of generative models in practical problem-solving contexts necessitates certain key characteristics. These include (i) the ability to produce high-quality samples, (ii) ensuring model coverage and diversity of samples, and (iii) guaranteeing quick execution times alongside computationally economical sampling (Xiao, Kreis, and Vahdat 2021). Generative models frequently strike a balance among different criteria. Specifically, GANs excel at rapidly generating high-quality samples, but they struggle with mode coverage and diversity of sampling (Wiatrak, Albrecht, and Nystrom 2019). In contrast, VAEs and normalizing flow models show proficiency in covering data modes but are often criticized for their low sample quality (Davidson et al. 2018; Asperti 2019).

GANs consist of two main components: a generator and a discriminator, also called a critic. They interact competently yet constructively, enhancing each other's performance. The generator's objective is to replicate real data distribution, while the discriminator, a type of binary classifier, assesses the probability that a specific sample is from the real dataset. The role of the discriminator is to improve its capacity to distinguish between synthetic and real samples. Nonetheless, GANs frequently encounter difficulties related to unstable training dynamics, for example, mode collapse, disappearing gradients, and problems with convergence (Wiatrak, Albrecht, and Nystrom 2019). This has led to a change in research direction towards creating more effective GAN variants (Miyato et al. 2018; Motwani and Parmar 2020). VAEs, on the other hand, maximize the Evidence Lower Bound (ELBO) to optimize the data's log-likelihood. Despite significant progress, VAEs still face challenges related to theoretical and practical aspects, such as balancing problems and variable collapse issues (Davidson et al. 2018; Asperti 2019). Generative models based on normalizing flows utilize a series of reversible transformations to convert a basic distribution into a more intricate one, achieving the desired probability distribution for the final target variable using the change of variables theorem. Unlike GANs and VAEs, these models explicitly learn the data distribution, and their loss function is simply the negative log-likelihood. However, these models have their limitations due to the necessity of a particular architecture for the Likelihood-based method to create a normalized probability model.

Additionally, for VAEs, an alternative loss like ELBO is not directly computed for the generated probability distribution. The GAN learning process is inherently unstable due to the adversarial nature of the GAN loss. Recently, diffusion models have emerged as potent generative models, becoming a central area of focus in computer vision, challenging researchers and practitioners to keep up with the fast-paced advancements (Sohl-Dickstein et al. 2015; Ho, Jain, and Abbeel 2020).

### 2.1. Diffusion Models:

Diffusion models are a resilient class of probabilistic generative models designed to decipher intricate data distributions. This is possible through a bifurcated process involving forward and reverse diffusion. In the forward diffusion phase, noise is incorporated into the input data, incrementally amplifying the noise level until it transforms into unadulterated Gaussian noise. This action alters the data distribution's architecture. Conversely, the reverse diffusion phase, also called denoising, is employed to restore the complex architecture of the original data from the altered and simpler data distribution. This action effectively reverses the deterioration instigated by the forward diffusion phase, culminating in a highly versatile and controllable generative model capable of precisely modeling intricate data distributions originating from random noise. The umbrella term 'Variational Perspective' includes models that leverage variational inference to approximate the desired distribution. This is commonly realized by minimizing the Kullback-Leibler divergence, a metric that quantifies the disparity between the approximate and desired distributions. A prime example of this class is the DDPMs. Initiated by (Sohl-Dickstein et al. 2015) and subsequently refined by (Ho, Jain, and Abbeel 2020), these models employ variational inference to approximate the parameters of a diffusion process (Sohl-Dickstein et al. 2015; Ho, Jain, and Abbeel 2020).

#### 2.1.1. Denoising Diffusion Probabilistic Models (DDPMs)

**The Forward Diffusion Process.** As outlined by DDPM, the forward diffusion process is depicted as a Markov Chain, characterized by the inclusion of Gaussian noise in a series of stages, culminating in the generation of noisy samples. It is important to note that the uncorrupted or original data distribution is represented as $q(x_0)$. With a data sample $x_0$ drawn from this distribution, $q(x_0)$, a forward noising operation, denoted as $p$, is employed. This operation introduces Gaussian noise iteratively at various time points, represented by $t$, resulting in a series of latent states $x_1$ through $x_T$. The process can be mathematically defined as follows:

$$q(x_t \mid x_{t-1}) = N\left(x_t : \sqrt{1-\beta_t}.x_{t-1}, \beta_t.I\right), \forall_t \in \{1, \dots, T\}, \tag{1}$$

In the context of this discussion, $T$ denotes the number of diffusion steps, while $\beta 1, \dots, \beta T$, each within the interval of [0, 1), signify the variance schedule spread throughout the diffusion steps. The identity matrix is symbolized by **I**, and $N(x;\ \mu,\ \sigma)$, characterizes the normal distribution possessing a mean of $\mu$ and a covariance of $\sigma$. By introducing $a_t$ as $1-\beta_t$ and defining $\underline{a_t} = \prod_{s=0}^{t} a_s$, it becomes feasible to directly sample from any step of the noised latent, conditioned on the initial input $x_0$ as follows:

$$q(X_t \mid X_0) = N\left(X_t : \sqrt{\underline{a_t}}.X_0, (1-\underline{a_t})I\right), \forall_t \in \{1, \dots, T\}, \tag{2}$$

$$X_t = \sqrt{\underline{a_t}}.X_0 + \sqrt{1-\underline{a_t}}\epsilon \quad \text{for} \quad \epsilon \sim N(0, I) \tag{3}$$

**Reverse Process.** Based on the definitions, we can construct an estimated inverse process for obtaining a sample from $q(x_0)$. This reverse process can be parameterized by initiating from $p(X_T) = N(X_T; 0, I)$, as detailed below:

$$p_\theta(X_{0:T}) = p(X_T) \prod_{t=1}^{T} p_\theta(X_{t-1} \mid X_t) \tag{4}$$

$$p_\theta(X_{t-1} \mid X_t) = N\left(X_{t-1}; \mu_\theta(X_t, t), \Sigma_\theta(X_t, t)\right) \tag{5}$$

The objective of training this model is to allow $p(x_0)$ to accurately learn and mimic the actual data distribution $q(x_0)$. This can be achieved by optimizing a specific variational bound associated with negative log-likelihood. This approach aligns with established methodologies and theories in the field (refer to sources).

$$E\left[-\log\log p_\theta(X_0)\right] \leq B_q\left[-\log\log \frac{p_\theta(X_{0:T})}{q(X_{1:T} \mid X_0)}\right] =$$

$$E_q\left[-\log\log p(X_T) - \sum_{t\geq 1} \log\log \frac{p_\theta(X_{t-1} \mid X_t)}{q(X_t \mid X_{t-1})}\right] = -L_{VL.B} \tag{6}$$

(Ho, Jain, and Abbeel 2020) proposed an innovative approach to parameterization. Instead of directly parameterizing $\mu_\theta(x_t, t)$ via a neural network, they developed a model $\epsilon_\theta(x_t, t)$ to predict the noise $\epsilon$ inserted during the forward process. This approach allowed them to reparametrize Equation (6), consequently simplifying the objective function as:

$$L_{simple} = E_{t,x_0,\epsilon}\left[ \|\epsilon - \epsilon_\theta(x_t, t)\|^2 \right] \tag{7}$$

where the authors draw a connection between the loss in Eq. (6) to generative score networks in (Song and Ermon 2019).

### 2.1.2. Score Perspective

The models based on scores employ a technique grounded in maximum likelihood estimation, using the data's log-likelihood score function to ascertain the diffusion process parameters. There are two main subsets in this category: Noise-conditioned Score Networks (NCSNs), as studied by (Song and Ermon 2019), and SDEs, researched by (Song et al. 2020). The primary objective of NCSNs is to calculate the derivative of the log density function for data distributions affected by noise at different levels. Conversely, SDEs serve as a broader approach, encompassing the characteristics of DDPMs and NCSNs. In the subsequent sections, a comprehensive analysis of the unique attributes of each subset is provided.

#### 2.1.2.1. Noise Conditioned Score Networks (NCSNs)

The score function of a given data distribution denoted as $p(x)$, is characterized as the gradient of the log density concerning the input, expressed mathematically as $\nabla_x \ log \, log \, p(x)$ . Intuitively, since the image $X_t$ comes from adding noise, the score function represent the direction that log-likehood function should follow to denoise the image. This score function can be approximated using a shared neural network trained via a process known as score matching. This involves the utilization of a score network, represented as $s_\theta$, a neural network model parameterized by $\theta$. The core aim of $s_\theta$ is to mimic the score of $p(x)$ (depicted as $p(x)(s_\theta(x) \approx \nabla_x \ log \, log \, p(x))$ by minimizing the corresponding objective:

$$E_{x\sim p(x)} \|s_\theta(x) - \nabla_x \ log \, log \, p(x) \|_2^2 \tag{8}$$

Computational complexity associated with calculating $\nabla_x \ log \, log \, p(x)$ tends to limit the scalability of score matching, particularly in deep networks and high-dimensional data. To address this issue, (Song and Ermon 2019) suggest the utilization of denoising score matching (Vincent 2011) and sliced score matching (Song et al. 2020). Moreover, (Song and Ermon 2019) underline significant obstacles that prevent a straightforward implementation of score-based generative modeling on real-world data. The primary issue is the inaccuracy of the estimated score functions in low-density regions since real-world data often cluster on low-dimensional manifolds in a high-dimensional space, aligning with the manifold hypothesis. To counter these challenges, the authors demonstrate the efficacy of data perturbation using Gaussian noise at varying scales. This process renders the data distribution more compatible with score-based generative modeling. A proposal is made to calculate the score corresponding to all noise levels via the training of a singular noise-conditioned score network (NCSN). The authors derive $\nabla_x \ log \, log \, p(x)$ as $\nabla_{xt} \, log \, log \, p_{\sigma_t}(x_t \mid x) = -\frac{x_t - x}{\sigma_t}$ , where the noise distribution is chosen to be $p_{\sigma_t}(x_t \mid x) = N(x_t : x, \sigma_t^2 . I)$ , with $x_t$ being a noisy version of $x$. Thus, given a sequence of Gaussian noise scales $\sigma_1 < \sigma_2 < \ldots < \sigma_T$ , the equation can be formulated as follows:

$$\frac{1}{T}\sum_{t=1}^{T} \quad \lambda(\sigma_T) E_{p(x)} \, E_{x_t \sim p_{\sigma_t}(x_t \mid x)} \|s_\theta(x_t, \sigma_t) + \frac{x_t - x}{\sigma_t} \|_2^2 \tag{9}$$

The weighting function is denoted as $\lambda(\sigma_t)$. To carry out the inference, an iterative method known as "Langevin dynamics" is utilized, which was originally discussed in the studies by (Parisi 1981) and later by (Grenander and Miller 1994). The Langevin dynamics constructs an MCMC procedure intended to procure samples from a specific distribution $p(x)$, utilizing its exclusive score function $\nabla_x \ log \, log \, p(x)$ .

In precise terms, the process starts by taking a random sample $x_0$ from the distribution $\pi(x)$. The subsequent step entails the iteration of a predefined sequence of operations to facilitate a gradual shift from $x_0$ towards other samples that follow the distribution $p(x)$.

$$x_i = x_{i-1} + \frac{\gamma}{2}\nabla_x \ log \ log \ p(x) \ + \ \sqrt{\gamma}.\omega_i \tag{10}$$

In the established process, $\omega_i$ is considered to follow a normal distribution, $N(0, I)$, where $i$ ranges from 1 to $N$. It is further observed that when $\gamma$ approaches 0 and $N$ tends towards infinity, the $x_i$ samples derived from this procedure ultimately converge to a sample from $p(x)$. Song and Ermon (2019) introduced a refinement to this algorithm, designating it as the "annealed Langevin dynamics" algorithm. This amendment was proposed to alleviate specific difficulties and failure points associated with score matching. In this amended algorithm, the noise scale $\sigma_i$ is progressively reduced or "annealed" over time (Song and Ermon 2019).

#### 2.1.2.2. Stochastic Differential Equations (SDEs):

Building upon the concepts introduced by the prior two methodologies, score-based generative models (SGMs) proposed by (Song et al. 2020), employ a unique transformation of the data distribution, $q(x_0)$, into noise. Notably, the distinguishing feature of SGMs is their ability to generalize the number of noise scales to infinity. This property allows us to conceptualize the previously discussed probabilistic models as discretized forms of an SGM. It is noteworthy that many stochastic processes, including the diffusion process, are found to be solutions to a specific type of stochastic differential equation (SDE), formulated as follows:

$$dX = f(X, t)dt + g(t)dW \tag{11}$$

The function $f(\cdot, t)$ signifies the drift coefficient of the Stochastic Differential Equation (SDE), while $g(t)$ symbolizes the diffusion coefficient. The variable $w$ is indicative of standard Brownian motion. Within this framework, $x_0$ represents the untainted data sample and $x_0$ stands for the disturbed data mirroring the standard Gaussian distribution. A unique characteristic of the SDE is a corresponding reverse time SDE, which functions in the opposite direction. Starting with a sample from $p_T$ and reversing this diffusion SDE process allows us to procure samples from our original data distribution $p_0$. The formulation for the reverse-time SDE is as follows:

$$dX = [f(X, t)dt - \ g^2(t)\nabla_x \ log \ log \ p_t(x) \ ] \ dt + g(t)d\underline{w} \tag{12}$$

The terms $dt$ and $\underline{w}$ represent the infinitesimally small negative time step and retrograde Brownian motion. To numerically resolve the reversed-time Stochastic Differential Equation (SDE), employing a neural network to approximate the score function is feasible. This approach, known as score matching, has been previously suggested and utilized in literature by (Song and Ermon 2019) and (Song et al. 2020). It allows for the estimation of $s_\theta(x, t)$ which can be represented as $\approx \nabla_x \ log \ log \ p_t(x)$ (highlighted in red in Equation (12)). This score model's training relies on a specific objective, which is given as follows:

$$L(\theta) = \ E_{X_t \sim p(X(t)| \ X(0)), X(0) \sim p_{data}} \left[\frac{\lambda(t)}{2} \| s_\theta(X(t), t) - \ \nabla_{x(t)} \ log \ log \ p_t(X(t) \mid X(0)) \ \|_2^2\right] \tag{13}$$

In the provided context, $\lambda$ represents a weighting function, while $t$ is drawn from a uniform distribution over the interval $[0, T]$. Notably, the term $\nabla_x \ log \ log \ p_t(x)$ is substituted by $\nabla_x \ log \ log \ p_{0t}(X(t) \mid X(0))$ to overcome certain technical challenges that may arise. It is also worth noting that the sampling process from SDEs can be efficiently executed by applying various numerical methods, as demonstrated in Equation (12). The subsequent sections comprehensively discuss three prevalent techniques typically employed in this context.

1. **Euler-Maruyama (EM) Approach:** This strategy employs a straightforward discretization method where the infinitesimal time increment, $dt$, is replaced with a finite time increment, $\Delta t$, and the infinitesimal Wiener process, $d\underline{w}$, is substituted with a Gaussian random variable, z, that follows a normal distribution with mean 0 and variance $\Delta t.I$. This allows for the resolution of Equation (12).

2. **Prediction-Correction (PC) Strategy:** This technique revolves around a nested loop of predictive and corrective procedures. The initial data is projected forward before being corrected in a series of steps. The Euler-Maruyama method can be used to solve the prediction aspect. As for the corrective phase, any score-based Markov Chain Monte Carlo (MCMC) procedure can be applied, including but not limited to annealed Langevin dynamics. Consequently, Langevin dynamics can be utilized to resolve Equation (10).

3. **Probability Flow Ordinary Differential Equation (ODE) Approach:** The SDEs represented in Equation (11) can be transcribed into ordinary differential equations according to the following method:

$$dX = \left[ f(X,t) - \frac{1}{2} g^2(t) \nabla_x \; log \; log \; p_t(x) \right] dt \qquad (14)$$

Therefore, through the resolution of the Ordinary Differential Equations (ODE) problem, the value of $x0$ can be obtained. However, despite its solution efficiency, ODE lacks a stochastic component necessary for error rectification, leading to a marginal compromise in its performance.

## 2.2. Transformers-based Models:

Language performs multiple roles as the foundation for human communication, cooperation, and education. It is a medium for articulating various emotions, encompassing affection, apprehension, and melancholy. Additionally, it aids in comprehending intricate notions such as justice, politics, and morality and proves essential in tackling complex problem-solving activities. Importantly, language is intrinsic to the creative arts, as illustrated by our extensive literary legacy and varied musical creations. Language significantly acts as a medium for transmitting and conservating knowledge across generations. Nevertheless, with the emergence of breakthroughs in AI, the Cartesian assertions about the exclusive human capacity for language are being challenged. The conventional human exclusive control over language is progressively diminishing, and AI is evolving to occupy this void. This scenario highlights the critical importance of LLMs, a groundbreaking progression in AI. To thoroughly understand the transformative capabilities of these models, it is essential to examine some key elements of related technologies, especially NLP.

Convolutional neural networks (CNNs) have typically been found to be less effective in NLP due to their inability to handle sequential data like text adequately. A more fitting alternative is the RNN, which has garnered widespread use due to its capacity to process such data. RNNs (Rumelhart, Hinton, and Williams 1985), specifically LSTM networks (Hochreiter and Schmidhuber 1997) and GRU networks (Chung et al. 2014), have been robustly recognized as leading-edge methodologies in sequence modeling and transduction challenges. This includes, but is not limited to, language modeling and machine translation (Bahdanau, Cho, and Bengio 2014; Cho et al. 2014; Sutskever, Vinyals, and Le 2014). In the aftermath of their development and successful implementation, a series of rigorous efforts have been launched to enhance further the capabilities of these recurrent language models and encoder-decoder architectures (Jozefowicz et al. 2016; Luong, Pham, and Manning 2015; Wu et al. 2016). Such endeavors underscore the growing potential and persistent refinement in this area of computational linguistics and AI. Recurrent architectures generally allocate computational processes according to the positional alignment of the input and output symbols within sequences. This implies that such models generate a series of hidden states, denoted as $h_t$, contingent on the prior hidden state $h_{t-1}$ and the input at position t. However, the sequential characteristic inherent in these models obstructs within-sample parallelization, becoming particularly impactful when managing lengthy sequences due to memory restrictions constraining batching across different examples. Nevertheless, recent research has demonstrated considerable enhancements in computational efficiency via factorization methodologies (Kuchaiev and Ginsburg 2017) and conditional computation techniques (Shazeer et al. 2017). These advancements have not only led to efficiency improvements but also enhanced model performance. Yet, the intrinsic limitation associated with sequential computation continues to persist.

Commencing in 2014, AI researchers embarked on explorations of new methodologies. Among these was the concept of 'attention', analyzing data sequences (such as text) and accentuating the most pertinent information while dismissing the remainder. This method proved instrumental in improving task accuracy. Despite its benefits, attention mechanisms were predominantly employed to boost the performance of RNNs, which implied that lingering performance issues remained unaddressed. However, a significant paradigm shift occurred in 2017. (Vaswani et al. 2017) proposed an innovative neural network architecture known as the 'transformer', transcending the capabilities of both CNNs and RNNs. After testing this architecture on eight GPUs and training the model for three and a half days, the researchers demonstrated promising results, particularly in English-to-French translations. During the period under discussion, there was an increasing recognition among scholars that conventional deep learning models were nearing their saturation point. This perspective hinged on the argument that large models yielded diminishing dividends.

However, the advent of the transformer model particularly galvanized the AI industry. At the heart of the transformer model lies the principle of self-attention, a concept that diverges considerably from a standard neural network. The training phase in self-attention involves simultaneous processing of all observations rather than employing a gradual progression. This model facilitates comprehension of the contextual essence of words. Notably, a transformer is typically a pre-trained model, indicating that the training of the dataset begins from a zero-knowledge state. The model does not benefit from any form of pre-existing knowledge. However, a pre-trained model is not immutable. It permits fine-tuning to accommodate modifications and enhancements, such as adding additional data. Consequently, we initiate our discussion by delineating the fundamental precept of the attention mechanism.

### 2.2.1. Self-attention:

Within the domain of a self-attention layer, as illustrated in Fig. 4, the initial operation involves the transformation of the input vector into three distinct vectors: the query vector (q), the key vector (k), and the value vector (v), each maintaining a constant dimensionality. These individual vectors are subsequently integrated into three distinct weight matrices $W^Q, W^K$ and $W^V$, respectively. A generalized representation of Q, K, and V can be derived from the input X:

$$K = W^K X, Q = W^Q X, V = W^V X, \tag{15}$$

$W^Q, W^K$ and $W^V$ signify mutable parameters within this context. The Scaled Dot-Product Attention mechanism is subsequently delineated as follows:

$$Attention\ (Q, K, V) = Softmax(\frac{QK^T}{\sqrt{d_k}})V, \tag{16}$$

The term $\sqrt{d_k}$ represents a scaling factor and a softmax function is utilized on the produced attention weights, effectively converting them into a mathematically normalized distribution.

### 2.2.2. Multi-Head Self-Attention:

The multi-head self-attention (MHSA) structure, as depicted in **Figure 4**, has been introduced by (Vaswani et al. 2017) as a method for modeling intricate relationships among token entities from diverse perspectives. The MHSA module particularly aids the model in simultaneously focusing on information from numerous representation sub-spaces, given the comparatively rough modeling proficiency of a single-head attention block. The functional mechanism of the MHSA can be theoretically outlined as follows:

$$MultiHead(Q, K, V) = [Concat(head_1, \dots, head_h)]W^O, \tag{17}$$

The term $head_i$ is representative of the Attention mechanism, where the mathematical expressions $Attention\ (QW_i^Q, KW_i^K, VW_i^V)$ are operational representations. Furthermore, $W^O$ denotes a linear mapping function applied for integrating the multi-headed representation. It is worth noting that h serves as a hyper-parameter, which, following the initial paper, was established at a value of h=8.

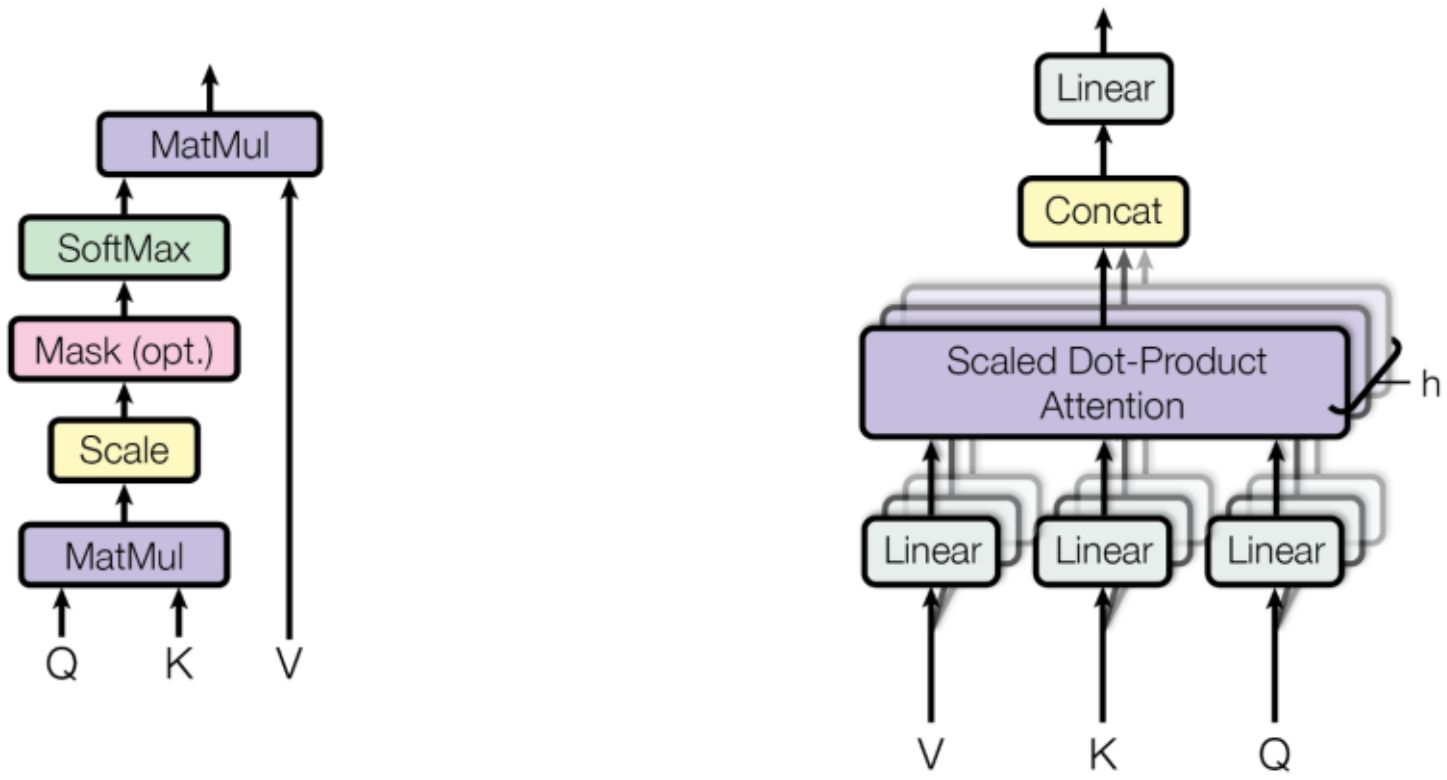

**Figure 4.** (left) We observe the representation of self-attention; (right) the illustration delineates the configuration of Multi-Head Attention, which comprises multiple concurrent attention layers (image by (Vaswani et al. 2017)).

The process under discussion involves an encoder utilizing a mechanism known as multi-head attention, which extends beyond the singular contextual comprehension found in self-attention. Specifically, multi-head attention can autonomously discern relationships among grammatical categories, such as nouns, adjectives, adverbs, and pronouns, without explicit programming by identifying emergent patterns. Upon completion of this phase, the resultant vectors are transferred to the decoder, undergoing a process analogous to the encoder's operation. A series of decoders process the data. In addition to multi-head attention, another technique, masked multi-head attention, is utilized. It functions by predicting subsequent words or tokens by considering the text preceding and following the masked location. It is noteworthy that the functions of encoders and decoders can be disentangled in the following ways: **(1) Encoder-only models**. These models are advantageous for text classification tasks such as sentiment analysis. An example of an LLM utilizing an encoder-only model is BERT. **(2) Decoder-only or autoregressive models**. These models are suitable for text generation tasks, akin to the predictive text functionality in a smartphone chat application. For instance, as you input text, the AI predicts the subsequent word or phrase. An example of this model is GPT-3. **(3) Encoder-decoder models**. These models facilitate generative AI tasks such as language translation and summarization. Notable LLMs employing this methodology include Facebook's BART and Google's T5.

### 2.2.3. Transformers in Computer Vision

Inspired by transformer-based models' successes in NLP, many research efforts have been made to develop their counterparts for computer vision tasks. Among them, ViT (Dosovitskiy et al. 2020) is the first attempt demonstrating that a solely transformer-based architecture can achieve superior performance in image recognition with pre-training on a substantial dataset. Recently, Swin Transformer (Z. Liu et al. 2021) attracted great attention due to its exceptional performance on several benchmarks for tasks such as image classification, object detection, and semantic segmentation. Compared to many previous transformer-based models, Swin Transformer proposed a hierarchical architecture whose representation is computed with shifted windows. This strategy enhances efficiency by restricting self-attention computation to non-overlapping local windows while also allowing for cross-window connection. A comprehensive overview of transformer-based models in medical imaging domain can be found in (Shamshad et al. 2023).

## 3. Various Applications of Deep Generative AI Models

### 3.1. Diffusion Models in Action:

Diffusion models, like other medical imaging methods, can be categorized in various ways. Here, we talk about ways to use diffusion in healthcare. These are broken down into five areas: (I) Image reconstruction, (II) Image-to-image translation, (III) Image generation, (IV) Image classification, (V) Other applications. **Figure 5** illustrates a compilation of studies within each category, detailing information like the specific algorithm used in the reverse process of the diffusion model for each study.

### 3.1.1. Image Reconstruction:

The essential role of medical image reconstruction in the domain of medical imaging is indisputable, especially in producing high-quality images for clinical use while concurrently striving to minimize expenses and risks to patients, as emphasized by (Levac, Jalal, and Tamir 2023) and (C. Cao et al. 2022). Widely used medical imaging techniques, such as Computerized Tomography (CT) and Magnetic Resonance Imaging (MRI), face challenges due to the fundamental physics involved, affecting their effectiveness, impeding their function, and potentially undermining the desired results. Increased radiation doses and extended tube occupancy times are necessary to obtain detailed, high-resolution patient data. However, these are only partially achievable due to health precautions and patient comfort considerations, underscoring the need for accelerated data collection in medical imaging methods like CT, PET, and MRI. Quicker imaging techniques can reduce examination times, enhance the availability of imaging services, and decrease waiting periods. They also improve the images' precision, particularly in dynamic studies that require rapid imaging sequences (Hyun et al. 2018; Korkmaz et al. 2021; C.-M. Feng et al. 2021). Consequently, strategies have been developed to decrease the radiation exposure from the standard dose or apply under-sampled or sparse-view imaging processes. Nevertheless, these approaches have limitations, such as reduced Signal-to-Noise Ratio (SNR) and Contrast-to-Noise Ratio (CNR). Addressing these obstacles and solving this ill-posed inversion issue is the responsibility of medical image reconstruction, as pointed out by (Gothwal, Tiwari, and Shivani 2022). The subsequent discussion will offer a synopsis of diffusion-based medical image reconstruction and enhancement frameworks.

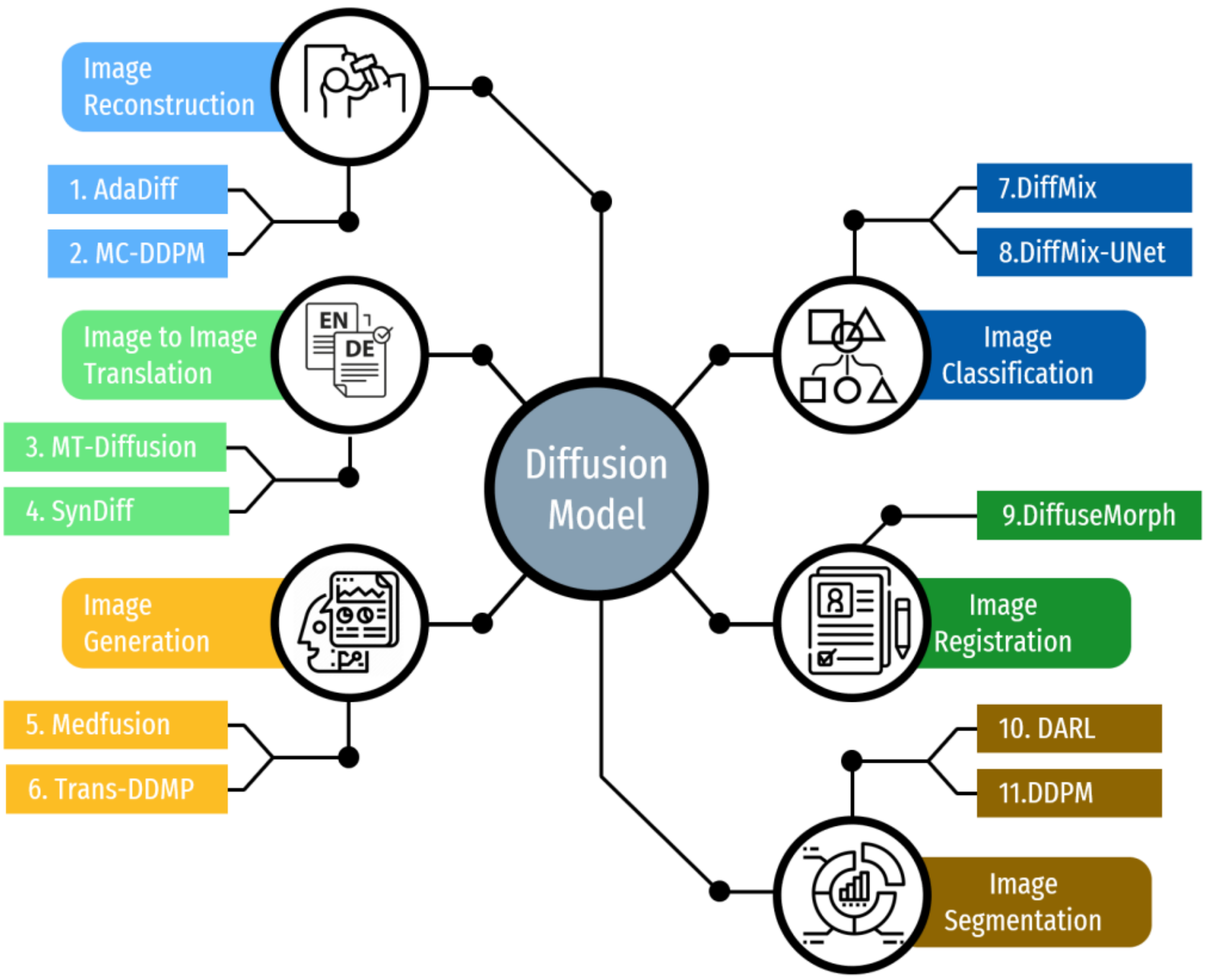

**Figure 5.** The proposed taxonomy for diffusion-based models in health care in six sub-fields, (I) Image Reconstruction 1. (Özbey et al. 2023), 2. (Xie and Li 2022), (Yang et al. 2024), (II) Image to Image Translation 3. (Lyu and Wang 2022), 4. (Özbey et al. 2023), (III) Image Generation 5. (Müller-Franzes et al. 2023), 6. (Pan et al. 2023), (Sun et al. 2024), (IV) Image Classification 7. (H.-J. Oh and Jeong 2023), 8. (Y. Yang et al. 2023), (V) Image Registraion 9. (Kim, Han, and Ye 2022), (VI) Image Segmentaion 10. (Kim, Oh, and Ye 2022), 11. (Azad et al. 2022), (Wu et al.2024).

In their recent study, (Özbey et al. 2023) introduced an innovative technique named Adaptive Diffusion Priors (AdaDiff) for the reconstruction of MRI. This approach involves a series of diffusion processes that enhance the authenticity of the generated images. Unlike traditional approaches utilizing static diffusion priors, AdaDiff dynamically adjusts its priors during the inference stage to align more closely with the distribution of the test data. The researchers demonstrated that this adjustment leads to superior results compared to existing methods regarding reconstruction quality and speed. Specifically, AdaDiff achieved a peak signal-to-noise ratio (PSNR) of 34.5 dB and a structural similarity index (SSIM) of 0.95 in a mere 1000 steps. Furthermore, the method demonstrated robustness against changes in the MRI acquisition protocol, maintaining consistent performance across various imaging operators **Figure 6**. Consequently, the authors deduced that AdaDiff holds great promise in enhancing the efficacy of MRI by facilitating swifter and more dependable image reconstruction.

In a recent study by (Xie and Li 2022), a new integrated approach termed a measurement-conditioned denoising diffusion probabilistic model (MC-DDPM) was introduced to reconstruct under-sampled medical images using the DDPM framework. This innovative technique demonstrated superior performance compared to existing methods in metrics such as PSNR and SSIM. It provided the ability to assess the uncertainty of the reconstruction. The researchers conducted a series of experiments using PD and PDFS datasets, which revealed that the MC-DDPM method surpassed other techniques by a considerable margin. Furthermore, they investigated the impact of varying the number of sampling steps and the total number of samples on the quality of the reconstructed images.

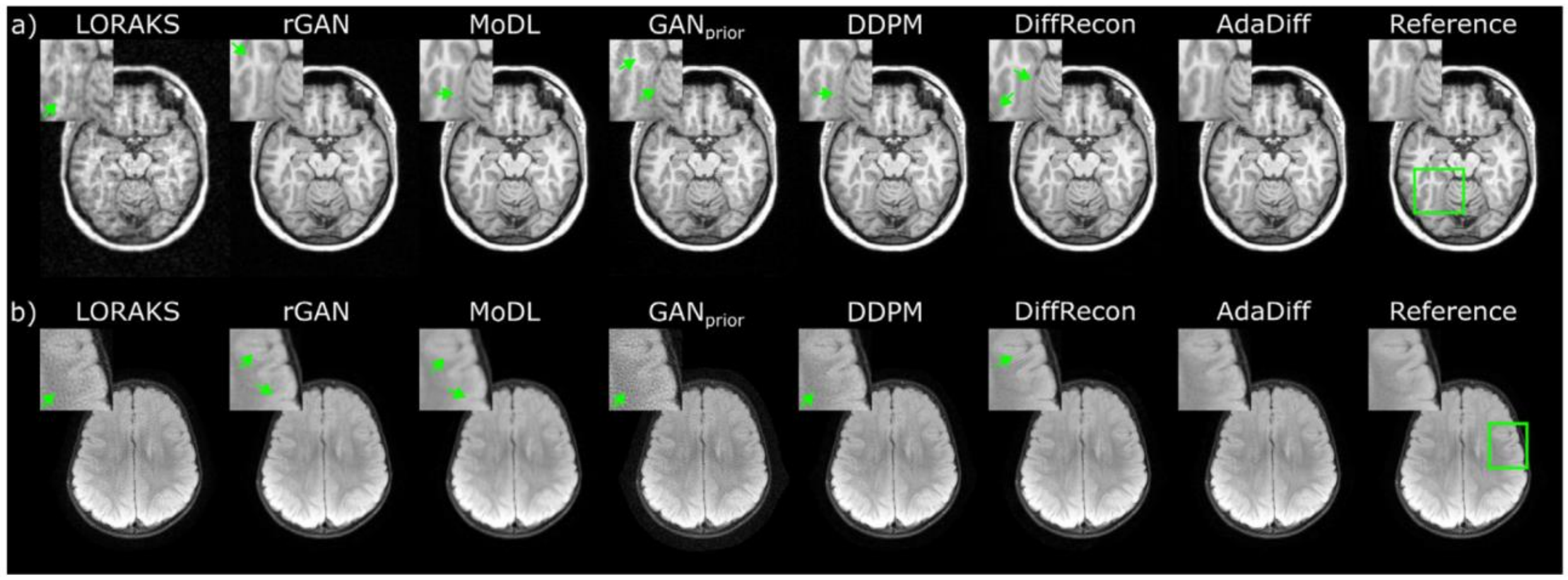

**Figure 6.** Results are shown for (a) T1-weighted acquisitions in the IXI dataset and (b) FLAIR-weighted acquisitions in the fastMRI dataset. Reconstructed images are given along with the reference image derived from fully sampled acquisitions, and zoom-in windows and arrows are included to highlight differences among methods. LORAKS and GAN prior show high noise amplification, rGAN shows residual aliasing, and MoDL shows visible spatial blurring despite high performance in quantitative metrics. DDPM has relatively higher noise among diffusion models, and DiffRecon shows local ringing artifacts near tissue boundaries. AdaDiff produces high-quality reconstructions with lower artifacts/noise and clearer tissue depiction than competing methods, images generated by (Özbey et al. 2023).

Interestingly, they discovered that a minor reduction in sampling steps only led to a slight decrease in PSNR. Additionally, they found that increasing the number of samples improved the quality of the sample mean and appeared to reach a plateau. Therefore, the authors suggested that 20 samples and 250 sampling steps might be an ideal selection for optimal efficiency. Moreover, the study also delved into the potential applications of MC-DDPM in areas beyond MRI reconstruction, including CT and PET reconstruction. In summary, the article proposes a promising technique for reconstructing under-sampled medical images, potentially enhancing the accuracy and efficiency of medical imaging procedures.

### 3.1.2. Image-To-Image Translation:

Recent advancements in image-to-image translation, a critical component in many applications, hinge on two fundamental architectures: pix2pix (Isola et al. 2017) and CycleGAN (Zhu et al. 2017; Shokrollahi et al. 2023). These systems have played a significant role in the medical field, particularly in diagnosis and therapy, where acquiring multi-modality images is crucial. There are instances, however, when certain modalities may be missing or unobtainable due to various conditions. To address this, diffusion models have been developed, demonstrating impressive capabilities in generating missing modalities by utilizing cross-modalities. An exemplary application of this technology can be seen in the translation from MRI to CT, enhancing the flexibility and completeness of medical imaging (Lyu and Wang 2022). CT and MRI are commonly used imaging techniques. However, CT struggles to capture detailed imagery of soft tissue injuries, often necessitating an MRI follow-up for a precise diagnosis. This two-step process can be lengthy and expensive and might lead to misalignment between the two sets of images.

To bridge this gap, harness the power of recently developed DDPMs and score-based diffusion models. Their approach includes a conditional DDPM and a conditional SDE, where the latter's reverse mechanism leverages T2w MRI images. The authors deploy DDPM and SDE in their research using three unique sampling methods: EM, PC, and ODE. When benchmarked against existing GAN (Gulrajani et al. 2017) and CNN (Ronneberger, Fischer, and Brox 2015) methodologies, their diffusion models show superior performance on the Gold Atlas male pelvis dataset (Nyholm et al. 2018). Key performance metrics include both SSIM and PSNR. Moreover, to gauge the reliability of diffusion models, (Lyu and Wang 2022) employ the Monte Carlo (MC) technique. Here, the model generates outputs ten times, with an average determining the conclusive result.

(Özbey et al. 2023) proposed a novel adversarial diffusion model called SynDiff for the unsupervised medical performance of medical image translation. SynDiff is designed to address the limitations of traditional GAN and diffusion models, which suffer from noise, spatial warping, and blurring. The authors conducted extensive experiments on multi-contrast MRI and MRI-CT translation and showed that SynDiff outperforms competing models regarding accuracy and artifact reduction **Figure 7**. The proposed method is based on a diffusion process that models the conditional distribution of the target image given the source image. The diffusion process is trained using a maximum likelihood objective, which is optimized using a stochastic gradient descent algorithm. The authors also introduce a novel adversarial loss function that encourages the generated images to be indistinguishable from the real images. The authors conducted several experiments to evaluate the performance of SynDiff on different medical imaging protocols. They compared SynDiff with several state-of-the-art GAN and diffusion models and showed that SynDiff outperforms these models regarding accuracy, artifact reduction, and computational efficiency. They also conducted a sensitivity analysis to evaluate the robustness of SynDiff to different hyperparameters and showed that SynDiff is relatively insensitive to these parameters. Overall, the proposed SynDiff method is a promising approach for unsupervised medical image translation that can improve the accuracy and efficiency of medical imaging protocols.

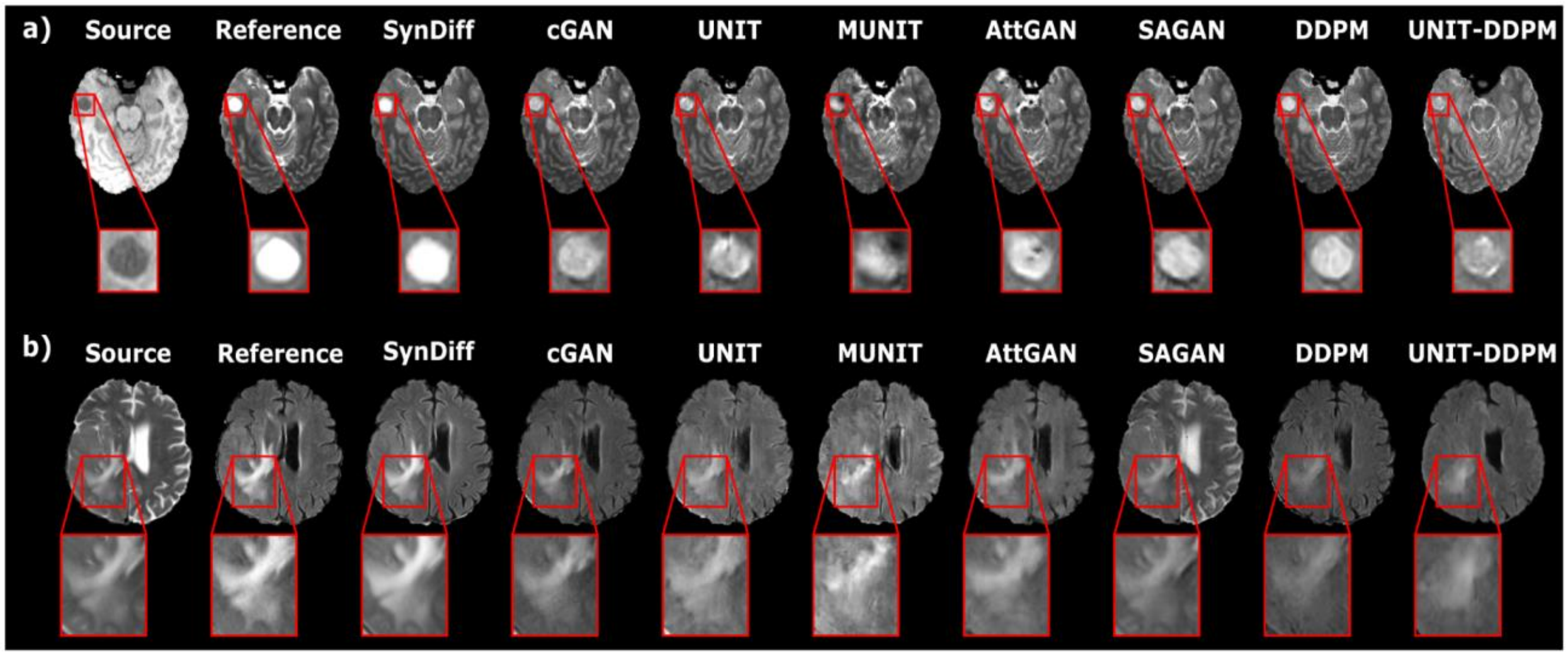

**Figure 7.** SynDiff showcased its capability for MRI contrast conversions on the BRATS dataset (Menze et al. 2014). For illustrative purposes, source images, synthesized outputs, and the actual reference images are presented for the following tasks: a) T1 to T2 and b) T2 to FLAIR (Fluid Attenuation Inversion Recovery). The visualization scales utilized are a) [0, 0.75] and b) [0, 0.80]. SynDiff effectively minimizes noise and artifacts, offering a clearer depiction of intricate structures than other baseline methods (Özbey et al. 2023).

### 3.1.3. Image Generation:

Diffusion models primarily focus on creating images and have been used extensively across different styles. Examples include the production of artificial 2D/3D medical visuals, as mentioned by (Pinaya et al. 2022), (Moghadam et al. 2023), (Dorjsembe, Odonchimed, and Xiao 2022), and (Kim and Ye 2022). Additionally, they have been utilized in converting 2D cell photographs into 3D representations, as cited by (Waibel et al. 2022). This segment delves into the techniques that use diffusion for generating medical imagery. (Müller-Franzes et al. 2023) Introduced Medfusion, a novel conditional latent DDPM for the generation of large clinical datasets. They assessed their diffusion model against several GANs across three datasets. These datasets comprised many RGB eye fundus images, chest CT scans, and 3D MRI brain scans from various subjects. The study involved training Medfusion on the 3D MRI brain scans and evaluating its effectiveness using several metrics, such as the PSNR, SSIM, and Fréchet inception distance (FID). The results indicated higher PSNR and SSIM values and lower FID scores for Medfusion, suggesting it outperformed GANs regarding image quality. Additionally, a sensitivity analysis was conducted to examine the influence of different hyperparameters on Medfusion's performance, revealing that some hyperparameters were more crucial than others. Ultimately, the researchers concluded that Medfusion is a promising technique for medical image synthesis, with advantages such as improved stability and interpretability over GANs.

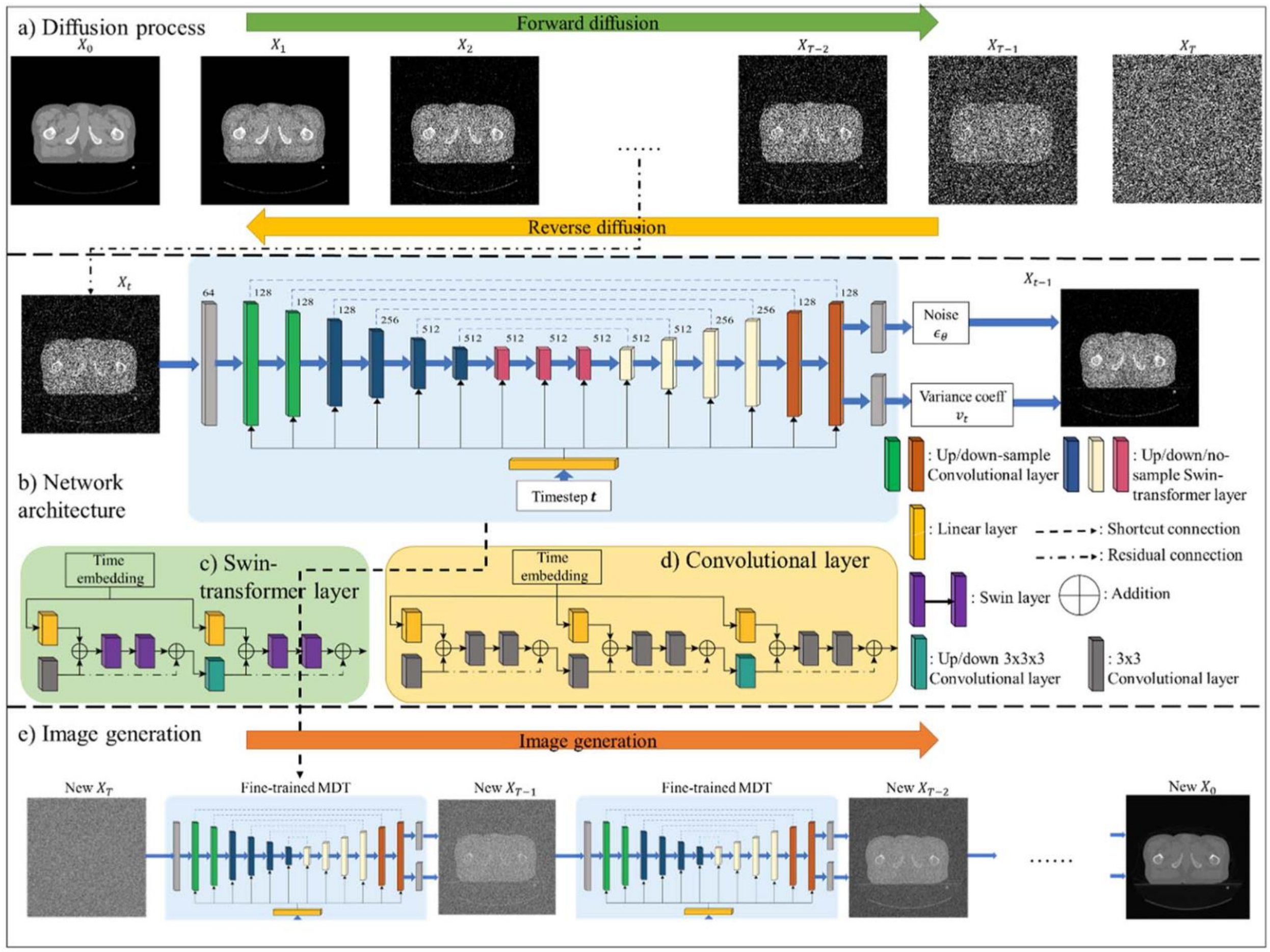

**Figure 8.** MT-DDPM Framework's Diffusion Methodology (a): Medical imagery is progressively transformed into pure Gaussian noise through incremental noise addition in the forward diffusion. For the backward process, a system is tasked to filter the Gaussian noise back into a pristine image continuously. (b): MT-DDPM Network Design: This network uses a balanced encoder-decoder design to master the backward process. The image devoid of noise is ascertained by forecasting the noise and its variance coefficient. (c): Component of Swin-transformer: The violet Swin segment comprises both window-based self-attention and a subsequent shifted window self-attention module. (d): Convolution Segment: This part houses three convolutional residual configurations that capture local image details. (e): Process of Image Creation: Upon advanced training, the MT-DDPM system can convert a fresh Gaussian noise image into a novel, synthesized image not previously found in the dataset (Pan et al. 2023).

Similarly, (Pan et al. 2023) developed a technique for synthesizing 2D medical images using a transformer-based denoising diffusion probabilistic model **Figure 8**. This innovative method employed a diffusion process modeled by a neural network trained to denoise images at each stage to generate samples from a probability distribution matching the target image distribution. The model, trained on a vast collection of medical images, was tested on several benchmark datasets using a transformer-based encoder-decoder network. A unique loss function was also introduced to encourage the generation of realistic and diverse samples. The authors conducted experiments on benchmark medical image datasets, including chest X-rays, heart MRIs, pelvic CT images, and abdomen CT images. They compared their approach with existing state-of-the-art methods. Their method outperformed the others in terms of image quality and fidelity. Using the Inception score and Fréchet Inception Distance score, quantitative evaluations confirmed that the synthetic images generated by their model belonged to the same data distribution as real images. Overall, this approach presents a promising avenue for enhancing diagnostic and treatment planning processes by creating high-quality medical images. The authors plan to extend this approach to 3D medical volume synthesis and explore more advanced network architectures to improve image synthesis quality further.

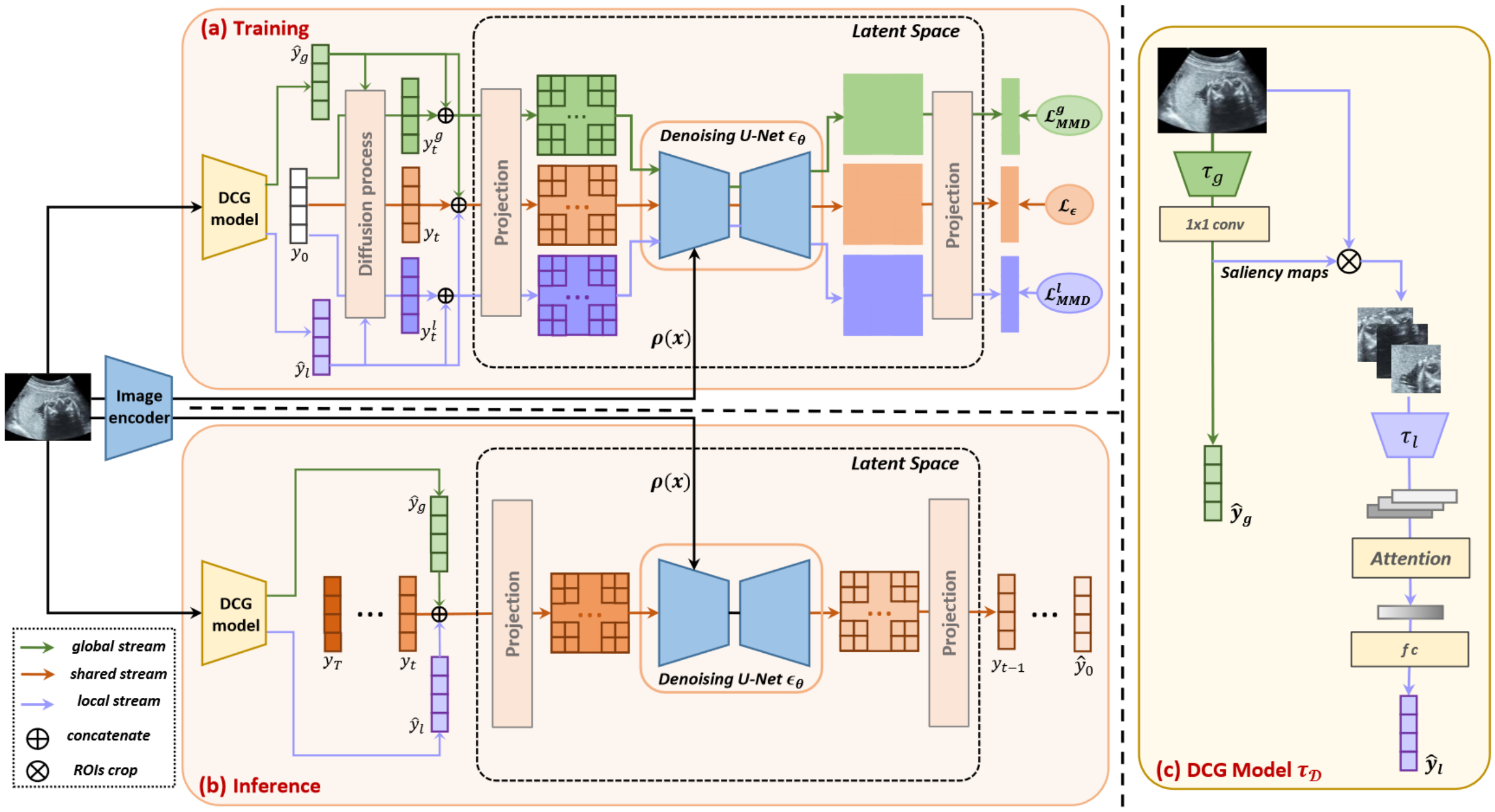

**Figure 9.** An overview of the DiffMIC framework includes (a) The forward process during the training phase (b) The reverse process for inference. (c) The DCG Model $\tau_D$ directs the diffusion using dual priors from the raw image and ROIs (Y. Yang et al. 2023).

### 3.1.4. Image Classification

The classification process is vital in the realm of medical image analysis. It paves the way for precise detection and description of various elements and irregularities found in medical imagery. This technique holds promise in transforming the medical sector by assisting healthcare experts in deciphering vast sets of intricate data (Shehab et al. 2022). However, there is still a pressing need to refine the integration of diffusion models for improved classification outcomes. In a recent study, (Y. Yang et al. 2023) proposed a DiffMIC method for classifying diverse medical imaging types via diffusion algorithms **Figure 9**. The process starts by translating the given image into a distinct feature space. The Dual-granularity Conditional Guidance (DCG) strategy is employed to gather overarching and minute details. After this, the diffusion of the actual image and the two prior sets produces a triad of noisy variables. These noisy entities are merged with their respective priors, taking them to a hidden layer resulting in three separate feature sets. To extract the noise-related characteristics of these sets, a denoising U-Net model is integrated with the image's original features. These refined feature sets are then reverted to their primal state.

In the Image Generation section, we discussed Diffusion models' potency in image synthesis. Recent advancements in machine learning (ML) have witnessed the emergence of state-of-the-art models such as GPT-3 (Brown et al. 2020b), DALL-E (Ramesh et al. 2022), Imagen (Saharia et al. 2022), and Stable Diffusion (Rombach et al. 2022), all of which have been fine-tuned on vast and varied internet datasets. Termed as "foundation models" (Bommasani et al. 2021), they epitomize exceptional generative capabilities, aiding in code development, producing art, and formulating text. In the medical domain, diffusion models have been harnessed to synthesize images, thereby augmenting datasets for tasks such as classification, among others. For example, In a recent study by (H.-J. Oh and Jeong 2023), the authors introduced a new data augmentation approach, 'DiffMix', specifically designed for segmenting and classifying nuclei in imbalanced pathology image datasets. This innovative framework leverages a semantic-label-conditioned diffusion model to create synthetic data samples, thereby enhancing the classification efficacy of rare nuclei types and exhibiting superior performance in both segmentation and classification tasks involving imbalanced datasets of pathology nuclei. A comprehensive evaluation of the proposed approach was conducted on two widely recognized imbalanced nuclei datasets, namely CoNSeP and MoNuSeg. The performance of DiffMix was benchmarked against well-established networks such as HoVer-Net and SONNET, as well as another data augmentation technique, GradMix. The findings from the experimental analysis revealed that DiffMix surpassed the other approaches with respect to classification accuracy, segmentation efficacy, and minimizing performance disparities across different class categories. Hence, the proposed approach holds significant potential to optimize the performance of various applications involving medical image processing tasks.

### 3.1.5. Other Applications of Diffusion Models:

Deformable **image registration** is a key method in medical imaging that finds flexible connections between two images that might move or change. It is important when the shape of images shifts because of reasons like the patient, when it is scanned, or the type of imaging used. (Kim, Han, and Ye 2022) presented a new method named DiffuseMorph. It consists of two primary systems: one for diffusion and one for deformation. Both systems are trained directly. The first system measures how the images change, and the second system estimates how they deform based on this measurement. Using spatial data, this method can smoothly show how one image transforms into another. **Image segmentation** is crucial in computer vision as it breaks down an image into important parts. This helps in medical studies by giving useful details about body structures. However, deep learning systems need varied data with detailed labeling to work well. There are limitations in the number of medical images and their labels due to the time, expense, and expertise needed (Azad et al. 2022). As a solution, diffusion models have become popular because they can create labeled data and reduce the need for detailed image annotations. (Kim, Oh, and Ye 2022) a new

DARL method was introduced for self-guided blood vessel segmentation to identify vascular disorders. This DARL approach consists of two primary components: one that learns the background image pattern and another that creates vessel segmentation visuals or artificial angiograms through a versatile SPADE technique (Park et al. 2019).

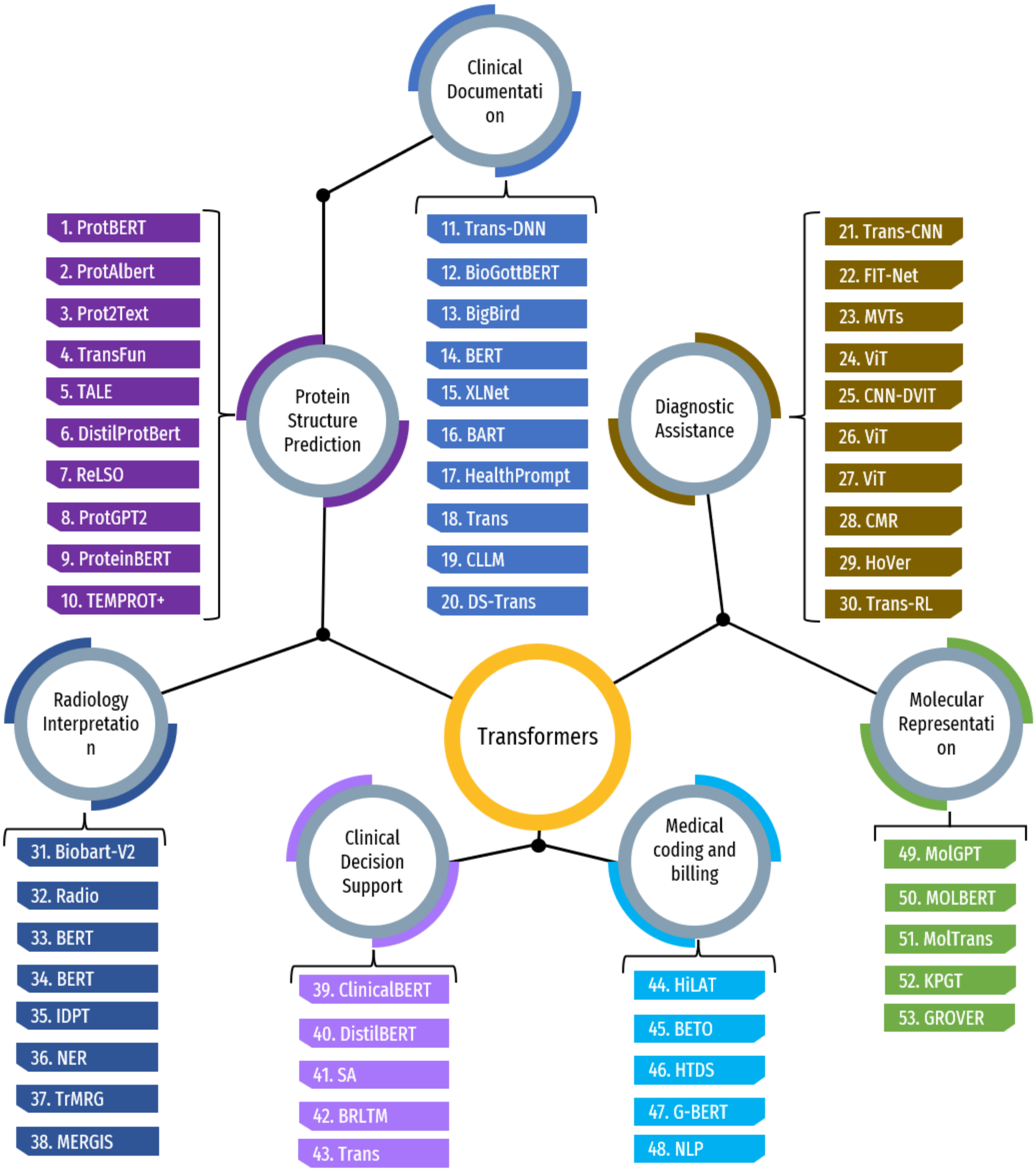

**Figure 10.** The proposed taxonomy for Transformer-based models in health care in seven sub-fields, (I) protein structure prediction 1. (Vig et al. 2020), 2. (Behjati et al. 2022), 3. (Abdine et al. 2023), 4. (Boadu, Cao, and Cheng 2023), 5. (Y. Cao and Shen 2021), 6. (Geffen, Ofran, and Unger 2022), 7. (Castro et al.

2022), 8. (Ferruz, Schmidt, and Höcker 2022), 9. (Ferruz, Schmidt, and Höcker 2022), (Zhang et al. 2024), 10. (Oliveira, Pedrini, and Dias 2023), (Abramson et al. 2024), (II) clinical documentation and information extraction 11. (Gérardin et al. 2023), 12. (Lentzen et al. 2022), 13. (Y. Li et al. 2022), 14. (Moon, He, and Liu 2022), 15. (S. H. Oh, Kang, and Lee 2022), 16. (Searle et al. 2023), 17. (Sivarajkumar and Wang 2022), 18. (Solarte-Pabón et al. 2023), 19. (Wei et al. 2022), 20. (Yogarajan et al. 2021) , (III) diagnostic assistance 21. (Azizi, Hier, and Wunsch Ii 2022), 22. (S. Chen et al. 2023), 23. (X. Chen et al. 2022), 24. (Dhinagar et al. 2023), 25. (Dong et al. 2023), 26. (Garaiman et al. 2023), 27. (Hosain et al. 2022), 28. (D. Hu et al. 2022), 29. (Mo et al. 2023), 30. (Zhou et al. 2023), (IV) medical imaging and radiology interpretation 31. (Balouch and Hussain 2023), 32. (Bhattacharya, Jain, and Prasanna 2022), 33. (Chaudhari et al. 2022), 34. (Jacenkow, O'Neil, and Tsaftaris 2022), 35. (J. Li et al. 2022), 36. (Moezzi et al. 2022), 37. (Mohsan et al. 2023), 38. (Nimalsiri et al. 2023), (V) clinical decision support 39. (J. Feng, Shaib, and Rudzicz 2020), 40. (W. Hu and Wang 2022), 41. (G. Huang 2022), 42. (Meng et al. 2021), 43. (Wang et al. 2023), (Renc et al. 2024), (VI) medical coding and billing 44. (L. Liu et al. 2022), 45. (López-García et al. 2023), 46. (Ng, Santos, and Rei 2023), 47. (Shang et al. 2019), 48. (Tchouka et al. 2023), (VII) drug design and molecular representation 49. (Bagal et al. 2022), 50. (Fabian et al. 2020), (Mendez-Lucio et al. 2024), 51. (K. Huang et al. 2021), 52. (H. Li, Zhao, and Zeng 2022), 53. (Rong et al. 2020).

## 3.2. Transformer-based Models in Actions:

Transformer-based models can be categorized in various ways. Here, we talk about ways to use transformers in healthcare. These are broken down into seven areas: (I) clinical documentation and information extraction, (II) medical coding and billing, (III) diagnostic assistance, (IV) medical imaging and radiology interpretation, (V) clinical decision support, (VI) protein structure prediction, (VII) drug design and molecular representation. **Figure 10** depicts a compilation of various studies for each category, detailing information like the particular algorithm employed in the transformer-based models.

### 3.2.1. Clinical Documentation and Information Extraction:

To enhance the effectiveness of subsequent NLP tasks, (Gérardin et al. 2023) devised and validated an algorithm for analyzing the layout of PDF clinical documents, extracting only clinically relevant text **Figure 11**. The algorithm involves several steps: text extraction using a PDF parser, text classification into categories like body text, left notes, and footers using a Transformer deep neural network architecture, and then aggregation of text lines associated with a given label. The algorithm was assessed technically and medically by applying it to a representative sample of annotated texts and examining the extraction of pertinent medical concepts. It was tested end-to-end in a medical use case involving automatic identification of acute infections detailed in a hospital report. (Gérardin et al. 2023) demonstrated that this preprocessing step improved the results in downstream tasks, such as medical concept extraction, proving its utility in a clinical context.

(Lentzen et al. 2022) thoroughly assessed the effectiveness of transformer-based AI models on German clinical notes. They employed a novel biomedical corpus for pre-training three new models. They systematically compared eight existing transformer-based models using a new dataset of clinical notes integrated with four other corpora. The newly trained BioGottBERT model outperformed the GottBERT model in clinical named entity recognition (NER) tasks. Additionally, (Y. Li et al. 2022) introduced two domain-specific language models, ClinicalLongformer and Clinical-BigBird, pre-trained on a large corpus of clinical text, improving several downstream clinical NLP tasks, such as question answering, NER, and document classification. (Moon, He, and Liu 2022) analyzed a corpus of 500,000 clinical texts for sub-language characteristics across different practice environments and document types for the ten most common clinical sections. They used BERT, fine-tuned for bio-clinical characteristics, to extract named entities for problem, test, and therapy concepts, and Sentence-BERT for rapid clustering. The analysis revealed narrow scopes in evaluation and discharge summary documents and similar cluster distributions in Family Medicine and Primary Care practice environments, suggesting comparable sub-languages. Conversely, Emergency Medicine exhibited a unique sub-language, characterized by higher phrase frequencies in discontinuous clusters.

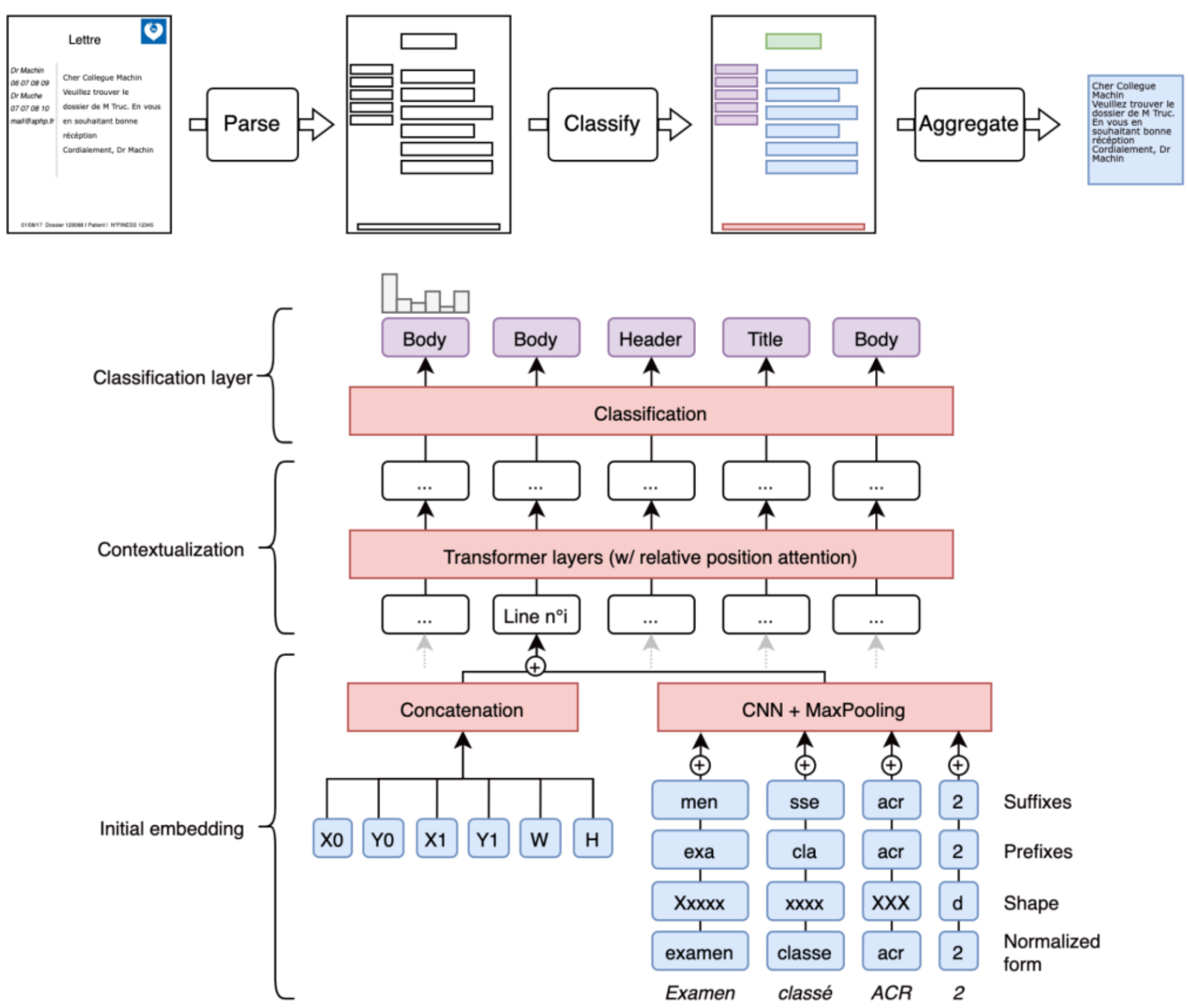

**Figure 11.** Structure of the Deep-Learning Line Classification Model. Textual and layout characteristics of each line are embedded to generate a unique representation for each line, which is subsequently contextualized using a four-layer Transformer that employs self-attention with relative position information. Finally, the representations are classified using a linear layer and a SoftMax function to determine the probability of each label (Gérardin et al. 2023).

(S. H. Oh, Kang, and Lee 2022) addressed the de-identification of protected health information (PHI) in medical documents and analyzed the efficacy of three prominent pre-training models (XLNet, RoBERTa, and BERT) for PHI recognition using the i2b2 2014 dataset. The dataset was tokenized and processed using an inside-outside-beginning (IOB) tagging approach and Word Piece tokenization. XLNet outperformed the other models in PHI recognition. Additionally, (Searle et al. 2023) explored summarization techniques for Biomedical Health Care (BHC) data and evaluated a new ensemble model that integrates extractive and abstractive summarization methods, using the SNOMED medical concept ontology as a clinical guidance signal. They proposed an advanced abstractive summarization model based on BART that includes a clinical oncology-aware guidance signal for key terms, facilitating the creation of problem-list-oriented abstractive summaries. (Sivarajkumar and Wang 2022) proposed a novel approach to clinical texts by implementing prompt-based learning, developing a groundbreaking prompt-based clinical

NLP framework called 'HealthPrompt'. This approach customizes task definitions via prompt template specification, eliminating the need for training data. (Solarte-Pabón et al. 2023) proposed a transformer-based approach for extracting named entities from Spanish clinical notes related to breast cancer, facilitated by a specialized breast cancer corpus and a schema for annotating clinical notes with breast cancer-related concepts. Their findings suggest that transformers are potentially effective in extracting information from Spanish clinical texts and can be enhanced by training in biomedical literature.

(Wei et al. 2022) introduced a multimodal technique called ClinicalLayoutLM for categorizing scanned clinical documents into specific categories, such as laboratory reports and CT scans. This technique combined text obtained from optical character recognition (OCR) with layout or image information. It outperformed the baseline model (which relied solely on OCR text) in classifying scanned clinical documents into 16 categories. (Yogarajan et al. 2021) considered transfer learning in the context of multi-label problem formulation for pre-trained models for predicting medical codes. They found that domain-specific transformers performed better for texts with fewer labels and/or documents shorter than 300 words. However, conventional neural networks were more effective for less frequently occurring labels and documents exceeding 300 words.

These studies highlight the utility of transformer-based models and other advanced techniques in various clinical NLP tasks, such as document classification, named entity recognition, and information extraction from clinical texts in multiple languages. However, the effectiveness of these models varies based on the task, language, and specific characteristics of the data, suggesting the need for a tailored approach to each application.

### 3.2.2. Medical Coding and Billing:

Medical coding and billing are a critical aspect of healthcare services, involving assigning standardized codes for illnesses and procedures in clinical documents. Recent advances in transformer-based models have shown promise in improving the efficiency and accuracy of this task. Several studies have proposed innovative approaches to facilitate the interpretable prediction of International Classification of Diseases (ICD) codes and to make the process more explainable. (L. Liu et al. 2022) The hierarchical Label-Wise Attention Transformer (HiLAT) model utilizes a two-tier hierarchical label-wise attention mechanism to generate label-specific document representations. These representations are then used to predict the likelihood of assigning a particular ICD code to a clinical document. The model was evaluated using the Medical Information Mart for Intensive Care III (MIMIC-III) database and outperformed previous state-of-the-art models for the 50 most frequently observed ICD-9 codes.

Similarly, (López-García et al. 2023) evaluated XLM-RoBERTa, mBERT, and BETO for explainable clinical coding, favoring a hierarchical task approach over multitask training, achieving superior results in medical-named entity recognition (MER) and medical-named entity normalization (MEN). (Ng, Santos, and Rei 2023) introduced Hierarchical Transformers for Document Sequences (HTDS), outperforming previous models for temporal modeling of document sequences in clinical notes. (Shang et al. 2019) proposed G-BERT, a model combining Graph Neural Networks (GNNs) and BERT, achieving superior medication recommendation accuracy by capturing hierarchical structures in medical codes. (Tchouka et al. 2023) developed a model for ICD10 code association, demonstrating its effectiveness as the most accurate solution for French clinical datasets to date, leveraging advancements in NLP and multi-label classification.

In conclusion, recent studies have demonstrated the effectiveness of transformer-based models in improving the efficiency and accuracy of medical coding and billing. These models leverage advanced NLP techniques, hierarchical attention mechanisms, and multi-label classification to process voluminous clinical documents and accurately assign ICD codes. At the same time, the models show promise in various languages and clinical contexts; ongoing research and development are necessary to optimize their performance further and make them more widely applicable in the healthcare industry.

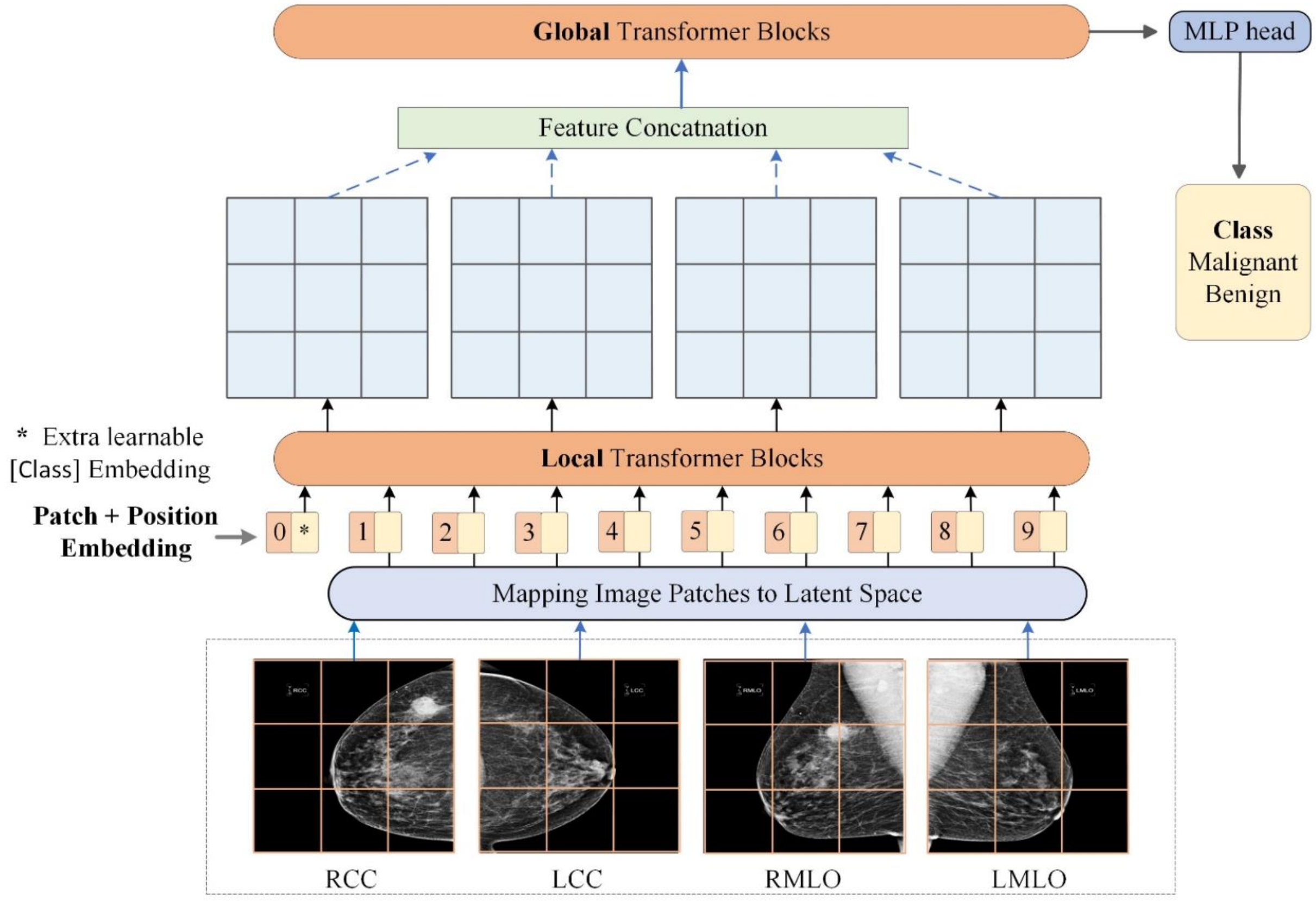

**Figure 12.** Structure of the Deep-Learning Line Classification Model. Textual and layout characteristics of each line are embedded to generate a unique representation for each line, which is subsequently contextualized using a four-layer Transformer that employs self-attention with relative position information. Finally, the representations are classified using a linear layer and a SoftMax function to determine the probability of each label (Gérardin et al. 2023).

### 3.2.3. Diagnostic Assistance:

Recent studies have showcased the potential of transformer-based models in various diagnostic applications. (Azizi, Hier, and Wunsch Ii 2022) developed a model for named entity identification, converting text segments with neurological signs and symptoms into clinical concepts using deep learning techniques. A comparative analysis between a transformer-based bidirectional encoder representations model and a CNNs model indicated the former's superiority in identifying signs and symptoms. This study is corroborated by (S. Chen et al. 2023), who introduced a feature interaction Transformer network (FIT-Net) for diagnosing pathologic myopia through Optical Coherence Tomography images, demonstrating superior performance over traditional deep learning methods. (X. Chen et al. 2022) leveraged Multi-view Vision Transformers (MVTs) for long-range correlations between mammograms, showcasing superior reproducibility, robustness, and case-based malignancy classification accuracy over other block combination alternatives.

In another study, (Dhinagar et al. 2023) investigated Vision Transformer (ViT) architecture iterations for high-stakes neuroimaging downstream tasks, focusing on gender and Alzheimer's disease classification using 3D brain magnetic resonance imaging. The study evaluated the effects of different ViT training strategies and emphasized their importance in neuroimaging applications. Similarly, (Dong et al.

2023) proposed an innovative end-to-end multi-label arrhythmia classification model, CNN-DVIT, combining a vision transformer structure with deformable attention and CNNs, outperforming the most recent transformer-based ECG classification methods. (Garaiman et al. 2023) aimed to develop a deep-learning model based on a vision transformer for identifying specific microangiopathy indicators in nail fold capillaroscopy images of patients diagnosed with Systemic Sclerosis. Although rheumatologists exhibited higher average accuracy, the ViT model was proficient in detecting various microangiopathic changes, highlighting the variability in performance among annotators **Figure 12**.

(Hosain et al. 2022) proposed an innovative approach for categorizing endoscopic imagery attributes using a vision transformer and transfer learning model, outperforming the DenseNet201 in identifying gastrointestinal diseases from wireless capsule endoscopy images. (D. Hu et al. 2022) introduced a cross-modal retrieval framework for executing retrieval tasks involving histopathology whole slide images and diagnostic reports, demonstrating the effectiveness of the proposed strategy in performing cross-modal retrieval tasks. (Mo et al. 2023) introduced an innovative region-of-interest-free model, HoVer-Transformer, for identifying breast cancer in ultrasound images, outperforming other CNN-based models and experienced sonographers in the interim analysis. Lastly, (Zhou et al. 2023) introduced a transformer-based representation-learning model for consolidating multimodal information processing as a clinical diagnostic tool, demonstrating superior performance in identifying lung illnesses and predicting adverse clinical outcomes in COVID-19 patients compared to image-only and non-unified multimodal diagnostic models.

In summary, transformer-based models have shown significant potential in various diagnostic applications, including named entity identification, pathologic myopia diagnosis, mammogram correlations, neuroimaging downstream tasks, arrhythmia classification, microangiopathy indicator identification, endoscopic imagery attribute categorization, histopathology image retrieval, breast cancer identification, and multimodal clinical diagnostic tool development. These studies collectively indicate the effectiveness and versatility of transformer-based models in diagnostic assistance, providing a foundation for future research and implementation in clinical settings to optimize clinical decision-making processes and patient triage.

### 3.2.4. Medical Imaging and Radiology Interpretation:

Medical imaging and radiology interpretation have seen significant advancements in recent years, with transformer-based models playing a crucial role in various aspects, such as summarization, disease diagnosis, error identification, image classification, information extraction, and report generation. Several recent studies have explored different aspects and applications of transformer-based models in medical imaging and radiology interpretation. One of the key challenges in radiology is generating pertinent summaries from the intricate and voluminous information in radiology reports. (Balouch and Hussain 2023) devised a special mechanism for abstractive text summarization using the Biobart-V2 model, which achieved state-of-the-art performance in summarizing medical texts, as evidenced by the ROUGE score. Additionally, (Bhattacharya, Jain, and Prasanna 2022) introduced the RadioTransformer. This groundbreaking framework assimilates radiologists' visual search patterns within a cascaded global-focal transformer framework, enhancing confidence in decision-making **Figure 13**.

Moreover, several studies have highlighted the importance of fine-tuning transformer models for specific tasks. (Chaudhari et al. 2022) developed a specialized version of BERT, Radiology BERT, fine-tuned artificially generated errors, substantially enhancing the identification of actual clinical speech recognition errors in radiology reports. Similarly, (Jacenkow, O'Neil, and Tsaftaris 2022) fine-tuned a BERT model for multimodal classification using dual image-text input, which improved image classification in radiology reports. Another study by (J. Li et al. 2022) utilized bidirectional encoder representations from the transformer (BERT)-based models and in-domain pre-training (IDPT), along with a sequence adaptation strategy, to develop an approach for the classification of actionable radiology reports in tinnitus patients, yielding promising results.

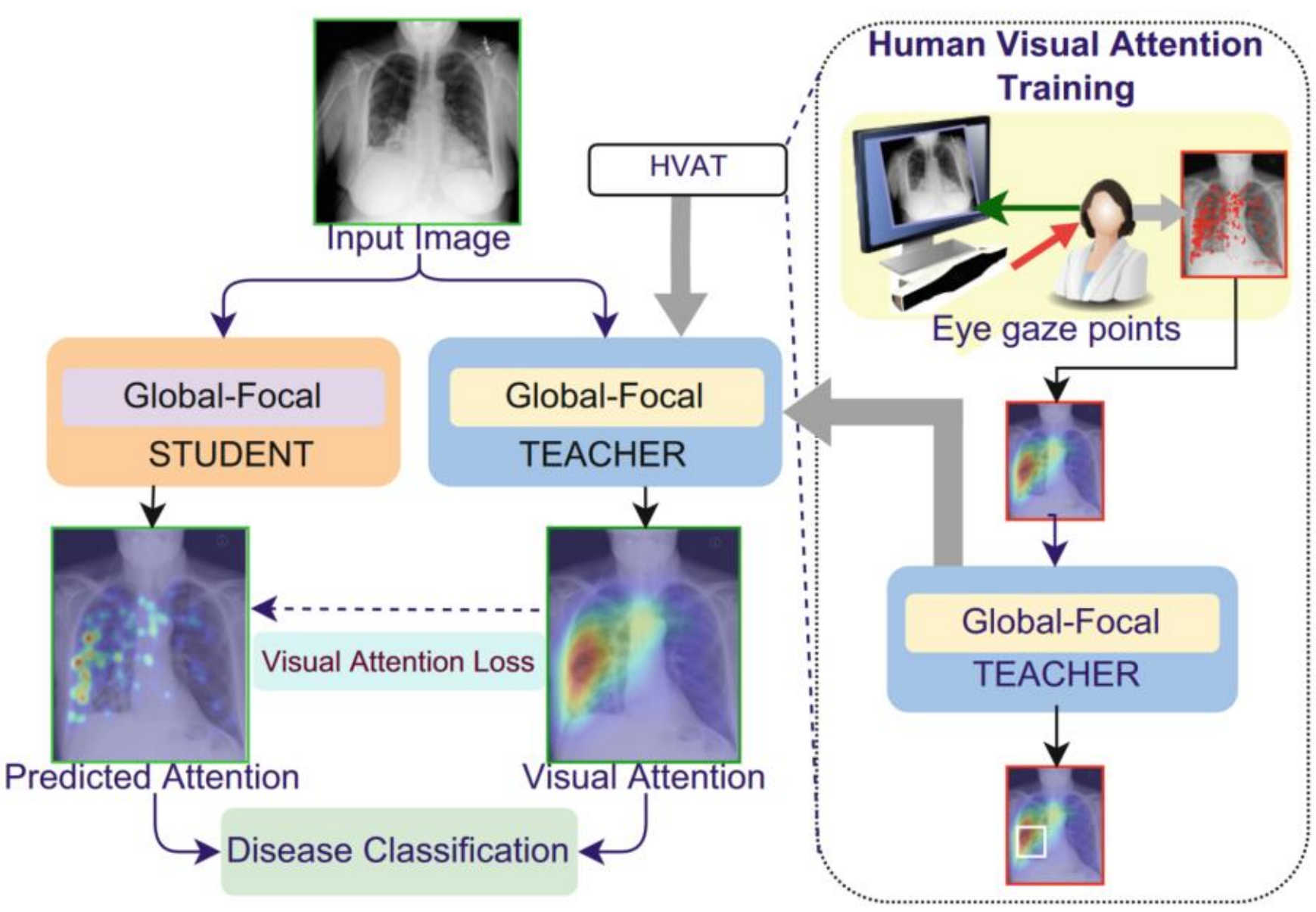

**Figure 13.** Schematic of the Proposed Methodology. The visual search patterns of radiologists on chest radiographs serve as the initial input for training a global-focal teacher network, denoted as Human Visual Attention Training. Subsequently, this pre-trained teacher network instructs the global-focal student network to acquire visual attention utilizing a newly devised Visual Attention Loss. Implementing the student-teacher network is meticulously designed to incorporate radiologist visual attention explicitly, thereby enhancing the accuracy of disease classification on chest radiographs (Bhattacharya, Jain, and Prasanna 2022).

Furthermore, in clinical information extraction, (Moezzi et al. 2022) employed a transformer-based fine-grained Named Entity Recognition (NER) architecture that surpassed the performance of previously utilized techniques, delivering a coherent and structured report. Also, (Mohsan et al. 2023) proposed a comprehensive transformer-based model, TrMRG, for report generation, which achieved noteworthy results compared to prevailing methods. Another novel approach, MERGIS, proposed by (Nimalsiri et al. 2023), utilized image segmentation and a modern transformer-based encoder-decoder model to enhance the accuracy of automated report generation. In conclusion, applying transformer-based models, often with task-specific fine-tuning, has shown promising results in various medical imaging and radiology interpretation aspects. These advancements in text summarization, disease diagnosis, error identification, image classification, information extraction, and report generation indicate that transformer-based models are vital tools for enhancing the accuracy and efficiency of radiology reports, ultimately contributing to better patient care.

### 3.2.5. Clinical Decision Support:

Transformer-based models have become increasingly prevalent in clinical decision support, as they hold significant potential for practical application in various aspects of clinical care. One novel approach involves the integration of a hierarchical CNN transformer with ClinicalBERT, a transformer-based model pre-trained on clinical corpora. This approach, proposed by (J. Feng, Shaib, and Rudzicz 2020), aims to capture the unique phrase-level patterns and global contextual linkages of medical language by leveraging the MIMIC-III database and investigating hidden attention layers. The study emphasizes the pragmatic utility of the model in sepsis prediction and ICU mortality by exclusively relying on textual data, exploring correlations between latent features extracted from structured and unstructured data, and recommending an evaluation framework for model efficacy. The findings, substantiated by domain expert evaluations, indicate that the model generates meaningful justifications that align with expert assessments, thus demonstrating its potential as a transparent decision-support tool.

Additionally, BERT-based models have been evaluated for their efficacy in predicting glaucoma progression using clinical free-text progress notes. A study by (W. Hu and Wang 2022) marked the initial efforts to apply BERT-based models, including the original BERT, BioBERT, RoBERTa, and DistilBERT, to ophthalmology clinical text, indicating potential in forecasting glaucoma progression. (G. Huang 2022) proposed a novel online surgical action detection and prediction model, termed SA, which leverages the attention mechanism of the Transformer to facilitate contextual decision-making in operating situations **Figure 14**. The SA model, an encoder-decoder framework, concurrently classifies current and subsequent surgical actions by ascertaining relationships between image components, surgical activities, and preceding images. Empirical results indicate superior performance in recognizing current actions compared to predicting future actions. Another innovative approach was proposed by (Meng et al. 2021), who developed a Bidirectional Representation Learning model (BRLTM) employing a Transformer architecture for multimodal Electronic Health Records (her) data. This model facilitates a two-stage approach of pretraining and finetuning fherEHR data modeling and enhances interpretability by quantitatively elucidating relationships between diffhernt EHR codes in sequences. The results indicate that bidirectional learning outperforms forward-only approaches in sequence modeling, thereby enabling the identification of individuals requiring depression screening and those at risk of developing depression within a specified time frame.

Lastly, (Wang et al. 2023) implemented a transformer-based approach in developing a family history (FH) lexical resource using a corpus of clinical notes from primary care settings. The study involved developing and evaluating rule-based and deep learning-based systems for FH information extraction. The integration of both systems enhanced the recall of FH information, although the F1 score remained variable yet comparable.

In summary, transformer-based models have shown significant potential in clinical decision support, spanning areas like sepsis and ICU mortality prediction, glaucoma progression forecasting, surgical action detection, depression diagnosis prediction, and family history information extraction. These approaches enhance interpretability, align with expert assessments, and improve prediction and classification of clinical outcomes. Despite challenges like variable F1 scores and better recognition of current actions than predicting future actions, these results suggest the promise of transformer-based models for practical use in clinical decision support.

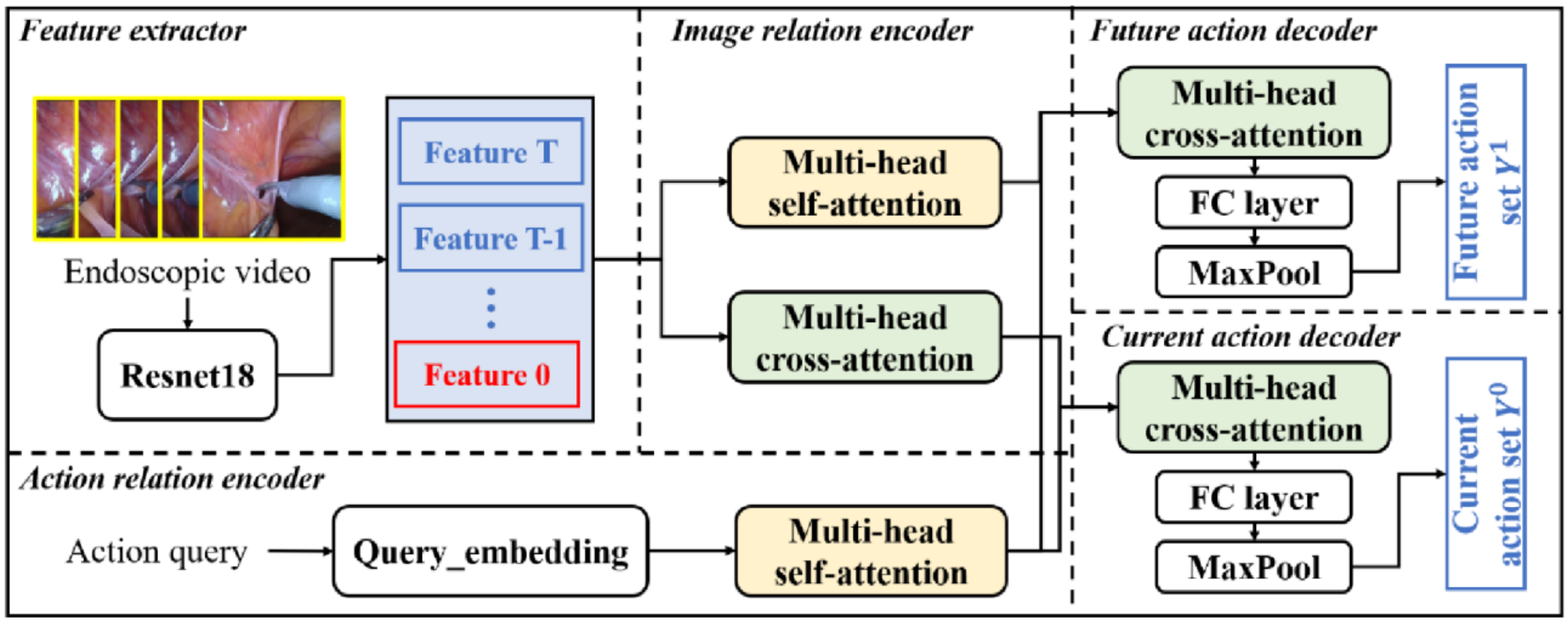

**Figure 14.** Depiction of the Proposed SA Model for Online Surgical Action Recognition and Prediction (G. Huang 2022).

### 3.2.6. Protein Structure Prediction:

Protein structure prediction is a critical area of research with significant implications for understanding protein functionality and drug design. Recent studies have employed transformer-based models, specifically designed for protein sequences, to predict various aspects of protein structure and function. (Vig et al. 2020) presented a comprehensive analysis of the interpretability of transformer models in the context of protein structures. They assessed five transformer models, including TapeBert, ProtTrans, ProtBert, ProtBert-BFD, ProtAlbert, and ProtXLNet, using attention mechanisms to elucidate proteins' structural and functional attributes. Their findings highlighted that attention mechanisms could reveal proteins' three-dimensional folding patterns and binding sites, which are crucial for their functionality. (Behjati et al. 2022) further developed this work by employing the ProtAlbert transformer and introducing novel methods for interpreting attention weights, leading to more accurate predictions of protein sequence profiles. In a different approach, (Abdine et al. 2023) proposed Prot2Text **Figure 15** This model generates textual descriptions of protein functions by combining Graph Neural Networks (GNNs) and LLMs, thus providing detailed and accurate descriptions of protein functions.

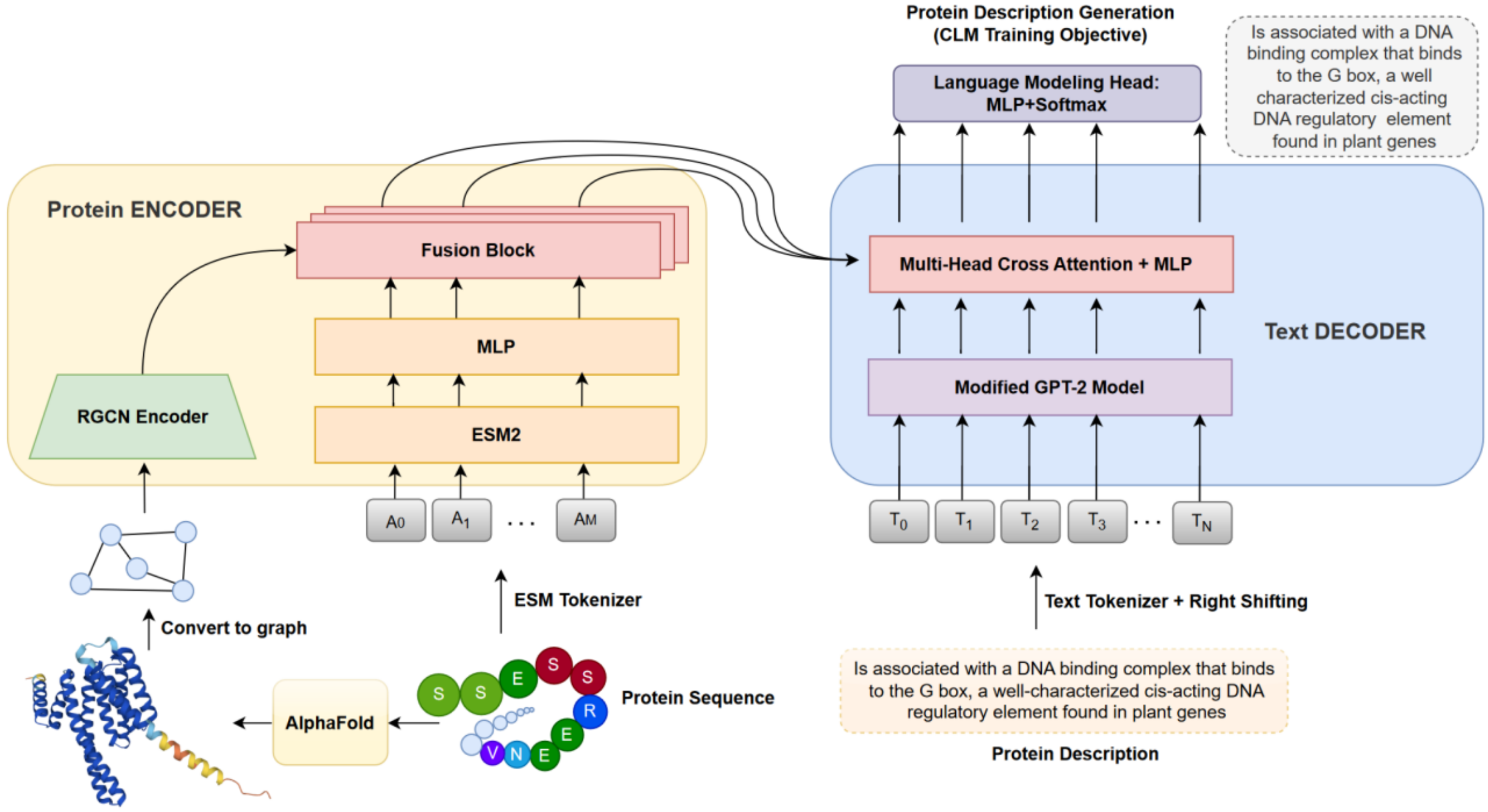

**Figure 15.** The proposed architecture of the Prot2Text framework is designed for predicting protein function descriptions in the free-form text (picture by (Abdine et al. 2023)).

Recent advancements in protein structure prediction and function annotation have been driven by innovative techniques such as TransFun (Boadu, Cao, and Cheng 2023), TALE (Y. Cao and Shen 2021), DistilProtBert (Geffen, Ofran, and Unger 2022), ReLSO (Castro et al. 2022), ProtGPT2 (Ferruz, Schmidt, and Höcker 2022), ProteinBERT , (Brandes et al. 2022), TEMPROT, and TEMPROT+ (Oliveira, Pedrini, and Dias 2023). TransFun combines protein sequences and 3D structures for accurate function prediction, while TALE utilizes deep learning and joint latent space embedding for improved generalization. DistilProtBert excels in distinguishing actual protein sequences while minimizing computational resources, and ReLSO optimizes protein sequences with a transformer-based autoencoder for fitness landscape navigation. ProtGPT2 generates novel protein sequences, while ProteinBERT integrates local and global representations for comprehensive processing. TEMPROT and TEMPROT+ fine-tune pre-trained architectures on protein sequences with BLASTp integration for enhanced performance. These advancements highlight the potential of transformer-based models and innovative techniques to provide detailed insights, ultimately advancing biomedical applications such as drug design.

### 3.2.7. Molecular Representation and Drug Design:

The recent advancements in transformer-based models have shown remarkable potential in molecular representation and drug design. These models leverage the capabilities of transformer decoders, masked self-attention, and self-supervised learning to generate valid, distinctive, and innovative compounds, predict drug-target interactions, and manage molecular properties. (Bagal et al. 2022) introduced MolGPT, a model designed to generate compounds with targeted scaffolds and chemical characteristics by utilizing scaffold SMILES strings. The model could construct molecules with property values that deviate from the provided values while maintaining the ability to produce molecules with user-specified scaffolds **Figure 16**. (Fabian et al. 2020) implemented a BERT-based model, MOLBERT, which surpassed the prevailing state-of-the-art models on benchmark datasets by utilizing learned molecular representations. The study underscored the importance of selecting appropriate self-supervised tasks during pre-training to enhance the dependability of the acquired representations and improve performance in downstream tasks such as Virtual Screening.

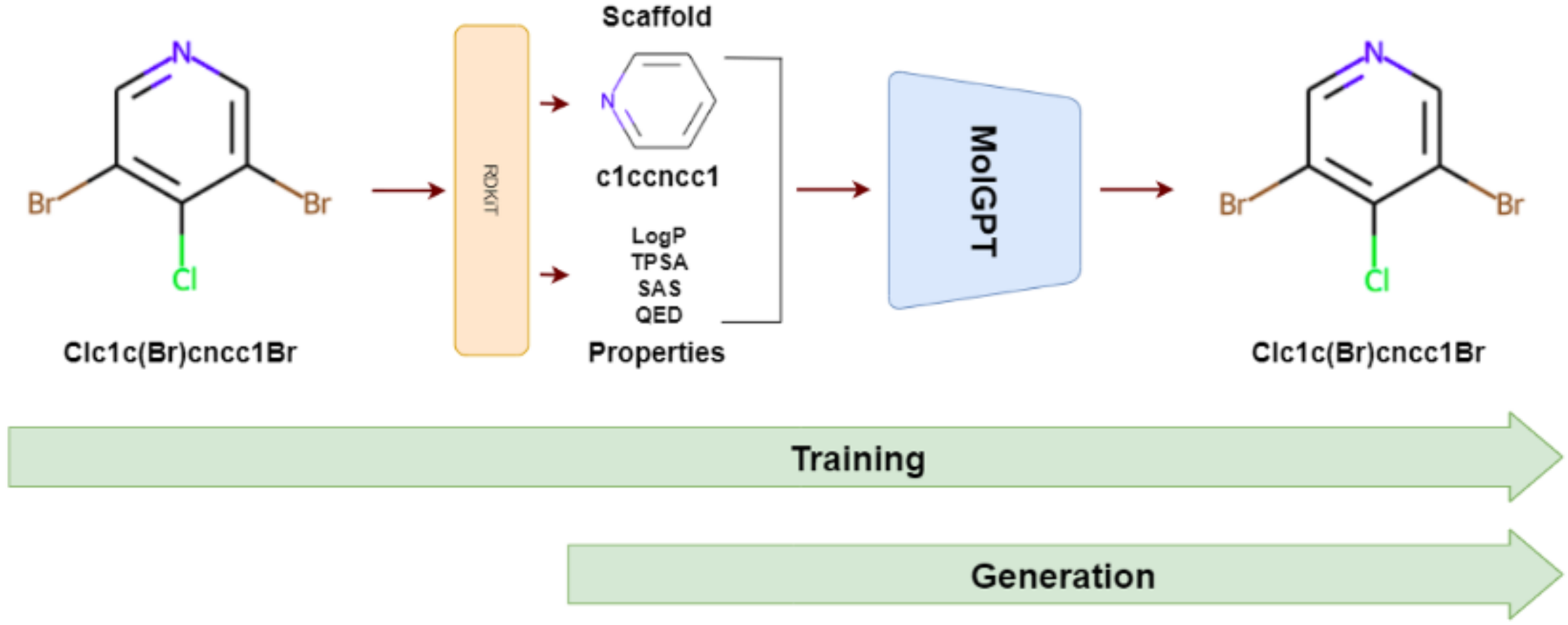

**Figure 16.** Pipeline for training and generation using the MolGPT model (Bagal et al. 2022).

(K. Huang et al. 2021) proposed MolTrans, an approach that combines a knowledge-inspired sub-structural pattern mining algorithm, an interaction modeling module, and an enhanced transformer encoder for more precise and interpretable drug-target interaction (DTI) predictions. This approach outperformed leading-edge baseline models in a comparative analysis using real-world data. (H. Li, Zhao, and Zeng 2022) introduced the KPGT framework, which utilizes a knowledge-guided pre-training approach to learn molecular graph representations and overcome the current limitations of self-supervised learning methods in molecular property prediction. This approach surpassed existing state-of-the-art self-supervised learning approaches, demonstrating the effectiveness of the KPGT architecture. (Rong et al. 2020) developed GROVER, which interprets extensive quantities of unlabeled molecular data to glean intricate structural and semantic details about molecules. GROVER combines Message Passing Networks with Transformer-style architecture to create more expressive encoders for complex information. It identifies semantic motifs in molecular networks and predicts their existence in a molecule using graph embeddings and integrating domain knowledge.

## 4. Comparative Review

**Table 1** comprehensively categorizes the assessed diffusion model papers by their application, key findings, and the employed or inspired algorithms, including DDPMs, NCSNs, and SDEs. It highlights each algorithm's fundamental concepts and objectives and the potential practical applications that can be explored and implemented in future studies based on the reviewed literature. The literature encompasses a spectrum of applications such as image reconstruction, image-to-image translation, image generation, image classification, image registration, and image segmentation. These studies underscore the efficacy of generative models in diverse medical imaging applications, from image reconstruction and translation to generation, classification, registration, and segmentation.

The table indicates that specific applications, such as image reconstruction, denoising, and generation, have garnered more attention than segmentation, text-to-image translation, and registration. This is mainly because diffusion model theory aligns with the demanded image processing tasks in the healthcare sector. Additionally, diffusion models can effectively model the complex interactions between signals and noise in data, resulting in more accurate image reconstruction. While diffusion models have potential across different tasks, some may necessitate further modifications for adaptation. For instance, text-to-image translation requires an auxiliary network with robust text encoding capabilities.

Regarding the transformers, some models have been integrated in various applications, from protein structure and function prediction to clinical documentation, diagnostic assistance, molecular representation, and drug design. The covered studies in this work exemplify the versatility and effectiveness of transformer-based models and architectures in addressing diverse challenges across various biomedicine domains.

**Table 1:** Summary of the evaluated diffusion models and transformer-based models in healthcare

| Application | Paper | Networks | Application | Key Findings |
| --- | --- | --- | --- | --- |
| Image Reconstruction | (Özbey et al. 2023) | AdaDiff | MRI Reconstruction | AdaDiff leads to superior results compared to existing methods |
| | (Xie and Li 2022) | MC-DDPM | Undersampled Medical Images Reconstruction | MC-DDPM is a promising technique for reconstructing under-sampled medical images |
| | (Siqi Wang et al 2024 | | Positron emission tomography (PET) image reconstruction | Proposed method offers a viable solution for improving PET image |
| Image To Image Translation | (Lyu and Wang 2022) | DDPM, SDE | Translation from MRI to CT | The approach includes a conditional DDPM and a conditional SDE. The model generates outputs ten times, with an average determining the conclusive result. The approach shows superior performance when benchmarked against existing GAN and CNN methodologies. The authors employ the Monte Carlo technique to gauge the reliability of the diffusion models. |
| | (Özbey et al. 2023) | SynDiff | Unsupervised medical performance of medical image translation | The proposed SynDiff method is based on a diffusion process and a novel adversarial loss function. The authors conducted extensive experiments and showed that SynDiff outperforms state-of-the-art GAN and diffusion models regarding accuracy, artifact reduction, and computational efficiency. |
| | | | | Sensitivity analysis showed that SynDiff is relatively insensitive to different hyperparameters. |
| | (Bottani, S et al. 2024) | U-Net | Translation of contrast-enhanced T1-weighted brain MRI to non-contrast-enhanced images | Models can effectively convert contrast-enhanced to non-contrast-enhanced images, facilitating the use of heterogeneous clinical data for reliable feature extraction |
| Image Generation | (Müller-Franzes et al. 2023) | Medfusion (DDPM) | Medical imagery: RGB eye fundus images, chest CT scans, 3D MRI brain scans | Medfusion outperformed GANs regarding image quality (higher PSNR and SSIM values, lower FID scores). Some hyperparameters were more crucial than others for Medfusion's performance. Medfusion has improved stability and interpretability over GANs. |
| | (Pan et al. 2023) | Transformer-based denoising diffusion probabilistic model | Synthesizing 2D medical images: chest X-rays, heart MRIs, pelvic CT images, abdomen CT images | The method outperformed existing state-of-the-art methods in image quality and fidelity. The Inception score and Fréchet Inception Distance score confirmed that the synthetic images generated belonged to the same data distribution as real images. |
| | (Kidder, B.L. 2024) | DreamBooth diffusion model | Generation of high-quality brain MRI scans, including images depicting low-grade gliomas, as well as contrast-enhanced spectral mammography (CESM) and chest and lung X-ray images | The study demonstrates that diffusion models can effectively synthesize diverse medical images, preserving patient anonymity and mitigating re-identification risks during data exchange for research purposes |
| Image Classification | (H.-J. Oh and Jeong 2023) | DiffMix | Segmenting and classifying nuclei in imbalanced pathology images | DiffMix is a semantic-label-conditioned diffusion model that creates synthetic data samples, enhancing the classification efficacy of rare nuclei types. |
| | (Y. Yang et al. 2023) | DiffMIC | Classifying medical imaging via diffusion algorithms | The DiffMIC method classifies medical images by translating them into distinct feature spaces and employing a denoising U-Net model. |
| Image Registration | (Kim, Han, and Ye 2022) | DiffuseMorph | Deformable Image Registration | Introduced a new method consisting of two primary systems: one for diffusion and one for deformation. This method can smoothly show how one image transforms into another. |
| | Truckenmueller, P., et al. 2024 | | Automated patient-to-image registration using a lightweight, table-mounted robotic alignment tool integrated with robotic cone-beam CT for intracranial biopsies and stereotactic electroencephalography | The integration of intraoperative cone-beam CT with a robotic alignment tool enables high precision and seamless workflow in neurosurgical procedures, potentially replacing traditional frame-based and frameless methods |
| | Liu, Y., et al. 2025 | ScaMorph | Deformable image registration across various medical | ScaMorph outperforms existing methods in 3D medical image registration, effectively |
| | | | imaging tasks, including brain MRI and liver CT registrations | handling diverse deformation complexities and task requirements |
| Image Segmentation | (Azad et al. 2022) | DDPM | Image Segmentation | Diffusion models can create labeled data and reduce the need for detailed image annotations. |
| | (Kim, Oh, and Ye 2022) | DARL | Blood Vessel Segmentation | Introduced a new DARL self-guided blood vessel segmentation method to identify vascular disorders. This approach creates vessel segmentation visuals or artificial angiograms. |
| Protein structure prediction | (Vig et al. 2020) | TapeBert, ProtTrans, ProtBert, ProtBert-BFD, ProtAlbert, ProtXLNet | Protein structure and function prediction | Attention mechanisms can reveal proteins' three-dimensional folding patterns and binding sites. |
| | (Behjati et al. 2022) | ProtAlbert | Protein sequence profile prediction | Novel methods for interpreting attention weights led to more accurate predictions of protein sequence profiles. |
| | (Abdine et al. 2023) | Prot2Text | Protein function description | Combining Graph Neural Networks (GNNs) and LLMs provides detailed and accurate descriptions of protein functions. |
| | (Boadu, Cao, and Cheng 2023) | TransFun | Protein function prediction | Combining protein sequences and 3D structures leads to accurate protein function prediction. |
| | (Y. Cao and Shen 2021) | TALE | Protein function prediction | Embedding protein function labels and sequence features into a joint latent space enhances the model's generalization ability to novel sequences and functions. |
| | (Geffen, Ofran, and Unger 2022) | DistilProtBert | Protein sequence distinction | The streamlined ProtBert model performed exceptionally well distinguishing actual protein sequences from random ones while minimizing computational resources. |
| | (Castro et al. 2022) | ReLSO | Protein sequence optimization | Transformer-based autoencoder optimizes protein sequences for fitness landscape navigation. |
| | Ferruz, Schmidt, and Höcker 2022 | ProtGPT2 | Novel protein sequence generation | Language models trained on protein sequences can generate novel proteins that mimic natural ones. |
| | (Ferruz, Schmidt, and Höcker 2022) | ProteinBERT | Protein sequence processing | A specialized deep language model amalgamates local and global representations for comprehensive end-to-end processing. |
| | (Oliveira, Pedrini, and Dias 2023) | TEMPROT, TEMPROT+ | Protein sequence embedding extraction | Fine-tuning and extracting embeddings from pre-trained architectures and integrating with BLASTp enhances performance. |
| | (Brandes et al. 2022) | ProteinBERT | Protein sequence processing | A specialized deep language model amalgamates local and global representations for comprehensive end-to-end processing. |
| Clinical documentation and | (Gérardin et al. 2023) | Transformer deep neural network | Analyzing the layout of PDF clinical documents | Developed and validated an algorithm for extracting clinically relevant text from PDF clinical documents. The algorithm improved the results in downstream tasks such as |
| information extraction | | | | medical concept extraction, proving its utility in a clinical context. |
| Diagnostic assistance | (Lentzen et al. 2022) | BioGottBERT | German clinical notes | Newly trained BioGottBERT model outperformed the GottBERT model in clinical named entity recognition (NER) tasks. |
| | (Y. Li et al. 2022) | ClinicalLongformer, Clinical-BigBird | Clinical text | Introduced two domain-specific language models pre-trained on a large corpus of clinical text, improving several downstream clinical NLP tasks such as question answering, NER, and document classification. |
| | (Moon, He, and Liu 2022) | BERT, Sentence-BERT | Clinical texts across different practice environments | Analyzed a corpus of 500,000 clinical texts and found narrow scopes in evaluation and discharge summary documents and similar cluster distributions in Family Medicine and Primary Care practice environments. Emergency Medicine exhibited a unique sublanguage. |
| | (S. H. Oh, Kang, and Lee 2022) | XLNet, RoBERTa, BERT | De-identification of protected health information (PHI) | XLNet outperformed the other models in PHI recognition using the i2b2 2014 dataset. |
| | (Searle et al. 2023) | BART | Summarization of Biomedical Health Care (BHC) data | Proposed an advanced abstractive summarization model based on BART that includes a clinical oncology-aware guidance signal for key terms, facilitating the creation of problem-list-oriented abstractive summaries. |
| | (Sivarajkumar and Wang 2022) | HealthPrompt | Clinical texts | Proposed a groundbreaking prompt-based clinical NLP framework called 'HealthPrompt' that customizes task definitions via prompt template specification, eliminating the need for training data. |
| | (Solarte-Pabón et al. 2023) | Transformer-based approach | Spanish clinical notes related to breast cancer | Proposed a transformer-based approach facilitated by a specialized breast cancer corpus and a schema for annotating clinical notes, suggesting that transformers are potentially effective in extracting information from Spanish clinical texts. |
| | (Wei et al. 2022) | ClinicalLayoutLM | Categorizing scanned clinical documents | Introduced a multimodal technique that combined text obtained from optical character recognition (OCR) with layout or image information, outperforming the baseline model (which relied solely on OCR text) in classifying scanned clinical documents into 16 categories. |
| | (Yogarajan et al. 2021) | Domain-specific transformers | Predicting medical codes | Found that domain-specific transformers performed better for texts with fewer labels and/or documents shorter than 300 words, while conventional neural networks were more effective for less frequently occurring labels and documents exceeding 300 words. |
| | (Simmons, A et al. 2024) | Large Language Models (LLMs) including GPT-3.5, GPT-4, Claude 2.1, Claude 3, Gemini | Extraction of ICD-10-CM codes from unstructured inpatient clinical notes | Current LLMs exhibit poor performance in accurately extracting ICD-10-CM codes compared to human coders, with GPT-4 achieving the highest percent agreement at 15.2% |

| | | | |
|---|---|---|---|
| | Advanced, and Llama 2-70b | | |
| (Zhang, Z. et al. 2025) | | Identification of underlying atrial fibrillation after ischemic stroke | The study demonstrates that AI-driven MRI analysis can effectively detect atrial fibrillation in post-stroke patients, offering a novel, non-invasive approach for identifying high-risk individuals and improving stroke management strategies |
| (Rui He et al. 2025) | YOLO | Utilization of AI to enhance real-time detection and segmentation of nasopharyngeal carcinoma (NPC) during endoscopic examinations | The study developed a deep learning-based model employing the YOLO (You Only Look Once) network, which demonstrated high performance in accurately identifying and segmenting NPC lesions in real-time during endoscopic procedures. This advancement aids clinicians in early diagnosis and precise localization of tumors, potentially improving patient outcomes |
| (Azizi, Hier, and Wunsch Ii 2022) | Transformer-based bidirectional encoder representations model, CNNs | Named entity identification of neurological signs and symptoms. | Superiority of transformer-based model in identifying signs and symptoms. |
| (S. Chen et al. 2023) | Feature Interaction Transformer network (FIT-Net) | Diagnosing pathologic myopia through Optical Coherence Tomography images. | Superior performance over traditional deep learning methods. |
| (X. Chen et al. 2022) | Multi-view Vision Transformers (MVTs) | Long-range correlations between mammograms. | Superior reproducibility, robustness, and case-based malignancy classification accuracy over other alternatives. |
| (Dhinagar et al. 2023) | Vision Transformer (ViT) architecture | Gender and Alzheimer's disease classification using 3D brain magnetic resonance imaging. | Importance of ViT training strategies in neuroimaging applications. |
| (Dong et al. 2023) | CNN-DVIT (combination of vision transformer structure with deformable attention and CNNs) | Multi-label arrhythmia classification. | Outperformed the most recent transformer-based ECG classification methods. |
| (Garaiman et al. 2023) | Vision Transformer (ViT) | Identifying specific microangiopathy indicators in nailfold capillaroscopy images of patients with Systemic Sclerosis. | Proficient in detecting various microangiopathic changes despite higher average accuracy by rheumatologists. |
| (Hosain et al. 2022) | Vision transformer and transfer learning model | Categorizing endoscopic imagery attributes for identifying gastrointestinal diseases from wireless capsule endoscopy images. | Outperformed DenseNet201 in identifying gastrointestinal diseases. |
| (D. Hu et al. 2022) | Cross-modal retrieval framework | Executing retrieval tasks involving histopathology whole | Effective strategy in performing cross-modal retrieval tasks. |

| | | | | |
|---|---|---|---|---|
| | | | slide images and diagnostic reports. | |
| | (Mo et al. 2023) | HoVer-Transformer | Identifying breast cancer in ultrasound images. | Outperformed other CNNs-based models and experienced sonographers in interim analysis. |
| | (Zhou et al. 2023) | Transformer-based representation-learning model | Consolidating multimodal information processing as a clinical diagnostic tool for identifying lung illnesses and predicting adverse clinical outcomes in COVID-19 patients. | Superior performance compared to image-only and non-unified multimodal diagnostic models. |
| | (Tu, T. 2024) | Articulate Medical Intelligence Explorer (AMIE) | Conversational diagnostic AI for simulating clinician-patient dialogues | AMIE demonstrated greater diagnostic accuracy and superior performance on multiple clinical axes compared to primary care physicians in a randomized, double-blind crossover study involving text-based consultations. |
| | (Sun, L et al. 2024) | ED-Copilot | Reducing emergency department wait times through AI-assisted diagnostic recommendations | ED-Copilot effectively personalized treatment recommendations based on patient severity, highlighting its potential as a diagnostic assistant to improve efficiency in emergency departments |
| Medical imaging and radiology interpretation | (Balouch and Hussain 2023) | Biobart-V2 | Summarizing medical texts | The ROUGE score showed a state-of-the-art performance in summarizing medical texts. |
| | (Bhattacharya, Jain, and Prasanna 2022) | RadioTransformer | Assimilating radiologists' visual search patterns | Enhanced confidence in decision-making within a cascaded global-focal transformer framework. |
| | (Chaudhari et al. 2022) | Radiology BERT | Identification of clinical speech recognition errors | Enhanced the identification of actual clinical speech recognition errors in radiology reports. |
| | (Jacenkow, O'Neil, and Tsaftaris 2022) | BERT model | Multimodal classification using dual image-text input | Improved image classification in radiology reports. |
| | (J. Li et al. 2022) | BERT-based models, IDPT, sequence adaptation strategy | Classification of actionable radiology reports in tinnitus patients | Yielded promising results. |
| | (Moezzi et al. 2022) | Transformer-based fine-grained NER architecture | Clinical information extraction | The performance of previously utilized techniques was surpassed, delivering a coherent and structured report. |
| | (Mohsan et al. 2023) | TrMRG | Report generation | Achieved noteworthy results compared to prevailing methods. |
| | (Nimalsiri et al. 2023) | MERGIS | Automated report generation | Utilized image segmentation and a modern transformer-based encoder-decoder model to enhance the accuracy of automated report generation. |
| Clinical Decision Support | (J. Feng, Shaib, and Rudzicz 2020) | Hierarchical CNN transformer, ClinicalBERT | Sepsis prediction, ICU mortality | The model captures phrase-level patterns and global contextual linkages of medical language, generating meaningful justifications that align with expert |
| | | | | assessments, demonstrating its potential as a transparent decision-support tool. |
| | (W. Hu and Wang 2022) | BERT, BioBERT, RoBERTa, DistilBERT | Predicting glaucoma progression | The models indicate potential in forecasting glaucoma progression. |
| | (G. Huang 2022) | SA model (encoder-decoder framework) | Surgical action detection and prediction | Superior performance in recognizing current actions compared to predicting future actions, facilitating contextual decision-making in operating theatres. |
| | (Meng et al. 2021) | BRLTM (Bidirectional Representation Learning model) | EHR data modeling, depression screening, and risk prediction | Bidirectional learning outperforms forward-only approaches in sequence modeling, enhancing interpretability by elucidating relationships between different EHR codes in sequences, enabling the identification of individuals requiring depression screening and those at risk of developing depression within a specified time frame. |
| | (Wang et al. 2023) | Transformer-based approach | Family history information extraction | The integration of rule-based and deep learning-based systems enhanced the recall of FH information, although the F1 score remained variable yet comparable. |
| | (Oniani, D et al. 2024) | Large Language Models (LLMs) enhanced with Clinical Practice Guidelines (CPGs) | Incorporation of CPGs into LLMs to improve Clinical Decision Support (CDS) | Integrating CPGs into LLMs enhances their performance in providing accurate recommendations for COVID-19 outpatient treatment, demonstrating potential for broader applications in CDS |
| | (Ferber, D. et al 2024) | Autonomous AI agents utilizing Large Language Models (LLMs) | Clinical decision-making in oncology | The study introduces an AI system where LLMs autonomously coordinate specialized medical AI tools, effectively enhancing clinical decision-making in oncology scenarios |
| Medical coding and billing | (L. Liu et al. 2022) | Hierarchical Label-Wise Attention Transformer (HiLAT) | Medical coding and billing | HiLAT model uses a two-tier hierarchical label-wise attention mechanism to generate label-specific document representations and predict the likelihood of assigning a particular ICD code to a clinical document. It outperformed previous state-of-the-art models for the 50 most frequently observed ICD-9 codes. |
| | (López-García et al. 2023) | Multilingual transformer-based models (XLM-RoBERTa, mBERT, BETO) | Explainable clinical coding | A hierarchical task approach for training the models, comprising medical-named entity recognition (MER) and medical-named entity normalization (MEN), yielded superior results compared to a multitask approach. |
| | (Ng, Santos, and Rei 2023) | Hierarchical Transformers for Document Sequences (HTDS) | Temporal modeling of document sequences | HTDS processes textual and metadata content from a series of documents and outperforms the previously established state-of-the-art model when the entire collection of clinical notes is used as input. |
| | (Shang et al. 2019) | G-BERT model | Articulating medical codes and recommending medications | G-BERT combines the strengths of Graph Neural Networks (GNNs) and BERT for articulating medical codes and recommending medications. It captured the underlying hierarchical structures in medical codes and outperformed all baseline models regarding prediction accuracy for medication recommendation tasks. |
| | (Tchouka et al. 2023) | A model leveraging NLP and multi-label classification | ICD10 code association | The model was rigorously evaluated on a French clinical dataset and demonstrated its efficacy as the most accurate solution for ICD10 code association in the French language to date. |
| Molecular Representation and Drug Design | (Bagal et al. 2022) | MolGPT | Generate compounds with targeted scaffolds and chemical characteristics | MolGPT utilizes scaffold SMILES strings to construct molecules with property values that deviate from the provided values while maintaining the ability to produce molecules with user-specified scaffolds. |
| | (Fabian et al. 2020) | MOLBERT | Predict drug-target interactions, manage molecular properties and Virtual Screening. | MOLBERT utilized learned molecular representations and outperformed prevailing state-of-the-art models on benchmark datasets. The study highlighted the importance of selecting appropriate self-supervised tasks during pre-training. |
| | (K. Huang et al. 2021) | MolTrans | More precise and interpretable drug-target interaction (DTI) predictions | MolTrans combines a knowledge-inspired sub-structural pattern mining algorithm, an interaction modeling module, and an enhanced transformer encoder. It outperformed leading-edge baseline models in a comparative analysis using real-world data. |
| | (H. Li, Zhao, and Zeng 2022) | KPGT | Learn molecular graph representations, molecular property prediction | KPGT utilizes a knowledge-guided pre-training approach to overcome the current limitations of self-supervised learning methods and surpasses existing state-of-the-art self-supervised learning approaches. |
| | (Rong et al. 2020) | GROVER | Interpreting structural and semantic details about molecules, predict the existence of semantic motifs in molecules. | GROVER combines Message Passing Networks with Transformer-style architecture to create more expressive encoders for complex information. It identifies semantic motifs in molecular networks and predicts their existence in a molecule using graph embeddings and integrating domain knowledge. |
| | (Yang, Y. et al. 2024) | MolEdit3D | Structure-based drug design through 3D molecular generation and optimization | MolEdit3D combines 3D molecular generation with optimization frameworks, employing a novel 3D graph editing model pre-trained on extensive 3D ligand data. This approach enhances the generation of molecules with favorable target-dependent and target-independent properties, achieving state-of-the-art performance across various evaluation metrics |
| | (Wang, J et al. 2024) | Token-Mol | Tokenized drug design utilizing large language models | Token-Mol encodes comprehensive molecular information, including 2D and 3D structures, into tokenized formats, transforming drug discovery tasks into probabilistic prediction problems. Through fine-tuning and reinforcement learning, Token-Mol demonstrates performance comparable to or surpassing existing task-specific methods in molecular generation and property prediction, improving regression task accuracy by approximately 30% compared to similar token-only methods |

## 5. Future Direction and Open Challenges

Generative AI, including diffusion models and transformer-based models, has showcased remarkable potential in the healthcare domain, particularly in medical imaging and disease diagnostics, by overcoming hurdles encountered by earlier models. It does not necessitate labeled data, making it a potent tool for numerous medical applications, and it excels in efficiently representing high-dimensional data like images. However, despite its merits, generative AI encounters challenges, such as slow generation processes, limited adaptability to certain data types, and the incapability of performing dimensionality reduction. This article intends to thoroughly assess recent medical research utilizing generative AI and classify studies to underscore its potential. We aim to emphasize the significance of generative AI in refining healthcare methodologies and pinpointing areas warranting further exploration.

Key challenges encompass investigating various medical imaging modalities, acquiring semantically meaningful data representations, enhancing architecture design, and addressing privacy concerns. Generative AI is apt for probing diverse modalities for specific downstream tasks. Although most extant studies concentrate on CT and MRI modalities, other modalities like ultrasound imaging could also be gained from generative AI. Despite VAEs and GANs being designed to retain and learn meaningful data representations, generative AI models struggle to form semantically meaningful data representations in their latent space (Yang et al. 2022; Ye et al. 2024). This issue is crucial as semantically meaningful representations facilitate superior image reconstructions and semantic interpolations. The network structure of generative AI models is a pivotal design choice influencing their capacity to comprehend intricate data relationships and yield high-quality outcomes. Recent research has delved into transformer models, but a comprehensive understanding of their capabilities and limitations is still lacking. Generative AI models can discern underlying probability distributions of datasets and generate novel data points, which is beneficial for intricate causal inference, discovery, and counterfactual generation tasks. However, privacy remains a major concern in the medical community (Carlini et al. 2023). AI image synthesis models, including generative AI, face scrutiny for potentially infringing copyright laws and jeopardizing training data privacy.

Federated learning combined with generative AI can establish a robust learning platform in the medical domain, addressing privacy concerns and enhancing the quality of learned models. Additionally, reinforcement learning can be employed to estimate the optimal inversion path for the inverse problem-solving of generative AI models, where conventional optimization methods become computationally prohibitive or unfeasible. Future research should concentrate on these challenges to maximize the potential of generative AI in healthcare. Addressing these challenges will enhance the capabilities of medical imaging techniques, improve patient care, and contribute to the exciting and rapidly evolving field of generative AI in healthcare. Moreover, integrating diffusion and transformer models holds promise for addressing current challenges and limitations in healthcare applications. While diffusion models are potent tools for simulating disease spread, comprehending brain networks, and predicting drug efficacy, they struggle with managing high-dimensional data and capturing complex interactions. Conversely, transformer models, rapidly adopted in healthcare for tasks like clinical report generation, medical image segmentation, and drug-drug interactions, face challenges associated with interpretability, environmental impact, computational costs, fairness and bias, AI alignment, and data privacy and sharing.

Interpretability is a significant concern as deep learning systems are often perceived as "black box" models, impeding the systemic acceptance of AI-aided diagnostics in the medical domain. Although transformers provide some transparency through attention weights visualization, the interpretation is often fragmented and not robust (Chefer, Gur, and Wolf 2021; Sun and Lio 2025). There have been efforts to enhance the interpretability of transformer models, such as proposing B-cos transformers and generating attention visualizations and cosine similarity between learned clinical diagnoses embeddings (Böhle, Fritz, and Schiele 2023; Nerella et al. 2024). However, there is still a need for novel techniques to enhance interpretability tailored towards healthcare AI.

The environmental impact of AI advancements has been substantial, with large-scale deep learning models generating significant carbon dioxide emissions. Initiatives have been undertaken to promote energy-efficient hardware and algorithms, such as the United Kingdom National Health Service's net-zero emissions goal by 2040. However, global initiatives are needed to make AI sustainable and accessible worldwide. Computational costs associated with the high parametric complexity of transformers and the necessity for vast datasets present challenges for healthcare settings that typically require lightweight models for real-time predictions with minimal maintenance costs (Strubell et al. 2019; Nerella et al. 2024). Techniques such as pruning, knowledge distillation, and quantization can provide more efficient model implementations for deployment within practical hardware constraints (Lagunas et al. 2021; Sun et al. 2019; Yao et al. 2022; Dantas et al. 2024). Fairness and bias in AI models can lead to unfavorable treatment of specific patient groups, often arising from the under-representation of certain populations in training datasets. Continuous monitoring and auditing for fairness and bias post-deployment are necessary to ensure equitable healthcare outcomes. AI alignment involves designing AI systems that align with human values and goals while minimizing unintended consequences and harmful outcomes (Moor et al. 2023, Shen, et al. 2024; Cross et al, 2024). It is especially critical in healthcare to ensure that large-scale foundation models are ethical, responsible, respectful of patient privacy, and not causing harm. A clear set of standards and guidelines is needed to establish the ethical use of AI in healthcare.

Addressing these challenges across modalities, representation learning, architectural design, privacy, interpretability, sustainability, and fairness will be essential to unlock the full potential of generative AI in healthcare. Integrating diffusion and transformer models, guided by principled design choices and ethical safeguards, offers a promising path toward trustworthy, sustainable, and clinically impactful AI in medicine.

## 6. Conclusion

In this study, we explored extensively the literature surrounding diffusion and transformed-based models, emphasizing their use in healthcare. We differentiated the diffusion models into three main categories: DDPMs, NCSNs, and SDEs, and elaborated on the attention mechanisms within transformers. Then we investigated diffusion models' roles in tasks including image reconstruction, image translation, generation, classification, etc. Additionally, we discussed transformer-based models' applications in protein structure prediction, clinical documentation, diagnostic assistance, radiology interpretation, clinical decision-making, medical billing, and drug design. We offered structured categorization and a broad overview of key methods for each use case. As a closing remark, we highlighted future research directions. While we shed light on the burgeoning application of diffusion and transformed-based models in healthcare, we recognize that the domain is still nascent and evolving. As these models rise in prominence and undergo further research, we hope our review can play a crucial guide for researchers aiming to leverage them. We aspire for our analysis to spark further interest and investigation into the capabilities of the generative AI models in healthcare.

## 7. Acknowledgement

XXXXXXX

## 8. Statements & Declarations

**Ethics Approval** Ethical approval was obtained from the ethics committee of XXXXX

**Competing Interest** The authors declare that they have no known competing financial interests or personal relationships that could have appeared to influence the work reported in this paper.

**Consent to Participate** Informed consent was obtained from all individual participants included in the study.